%% file: main.tex
\documentclass[10pt,twocolumn,letterpaper]{article}

\usepackage{cvpr}              

\input{preamble}

\definecolor{cvprblue}{rgb}{0.21,0.49,0.74}
\usepackage[pagebackref,breaklinks,colorlinks,citecolor=cvprblue]{hyperref}
\def\paperID{5395} 
\def\confName{CVPR}
\def\confYear{2024}
\usepackage{makecell,pifont,multirow,multicol}
\title{HoloVIC: Large-scale Dataset and Benchmark for Multi-Sensor Holographic Intersection and Vehicle-Infrastructure Cooperative}
\vspace{-0.4cm}

\author{Cong Ma$^1$,\quad Lei Qiao$^1$,\quad Chengkai Zhu$^1$,\quad Kai Liu$^1$,\quad Zelong Kong$^1$,\\ Qing Li$^1$,\quad Xueqi Zhou$^1$,\quad Yuheng Kan$^1$,\quad Wei Wu$^{1,2}$\thanks{Corresponding Author}\\
$^1$SenseAuto Research\qquad $^2$Tsinghua University\\
\href{https://holovic.net}{https://holovic.net}
\vspace{-0.4cm}}

\begin{document}
\maketitle
\input{sec/0_abstract}    
\input{sec/1_intro}
\input{sec/2_formatting}

\input{sec/3_finalcopy}
\input{sec/4_Tasks}

\input{sec/5_experiments}

{
    \small
    \bibliographystyle{ieeenat_fullname}
    \bibliography{main}
}

\input{sec/X_suppl.tex}

\end{document}

%% file: preamble.tex
\usepackage[dvipsnames]{xcolor}


%% file: sec/0_abstract.tex
\begin{abstract}
Vehicle-to-everything (V2X) is a popular topic in the field of Autonomous Driving in recent years. Vehicle-infrastructure cooperation (VIC) becomes one of the important research area. Due to the complexity of traffic conditions such as blind spots and occlusion, it greatly limits the perception capabilities of single-view roadside sensing systems. To further enhance the accuracy of roadside perception and provide better information to the vehicle side, in this paper, we constructed holographic intersections with various layouts to build a large-scale multi-sensor holographic vehicle-infrastructure cooperation dataset, called HoloVIC. Our dataset includes 3 different types of sensors (Camera, Lidar, Fisheye) and employs 4 sensor-layouts based on the different intersections. Each intersection is equipped with 6-18 sensors to capture synchronous data. While autonomous vehicles pass through these intersections for collecting VIC data. HoloVIC contains in total on 100k+ synchronous frames from different sensors. Additionally, we annotated 3D bounding boxes based on Camera, Fisheye, and Lidar. We also associate the IDs of the same objects across different devices and consecutive frames in sequence. Based on HoloVIC, we formulated four tasks to facilitate the development of related research. We also provide benchmarks for these tasks. 


\end{abstract}

%% file: sec/1_intro.tex
\vspace{-0.3cm}
\section{Introduction}
\label{sec:intro}

\begin{figure}
	\centering
	\includegraphics[width=8.2cm]{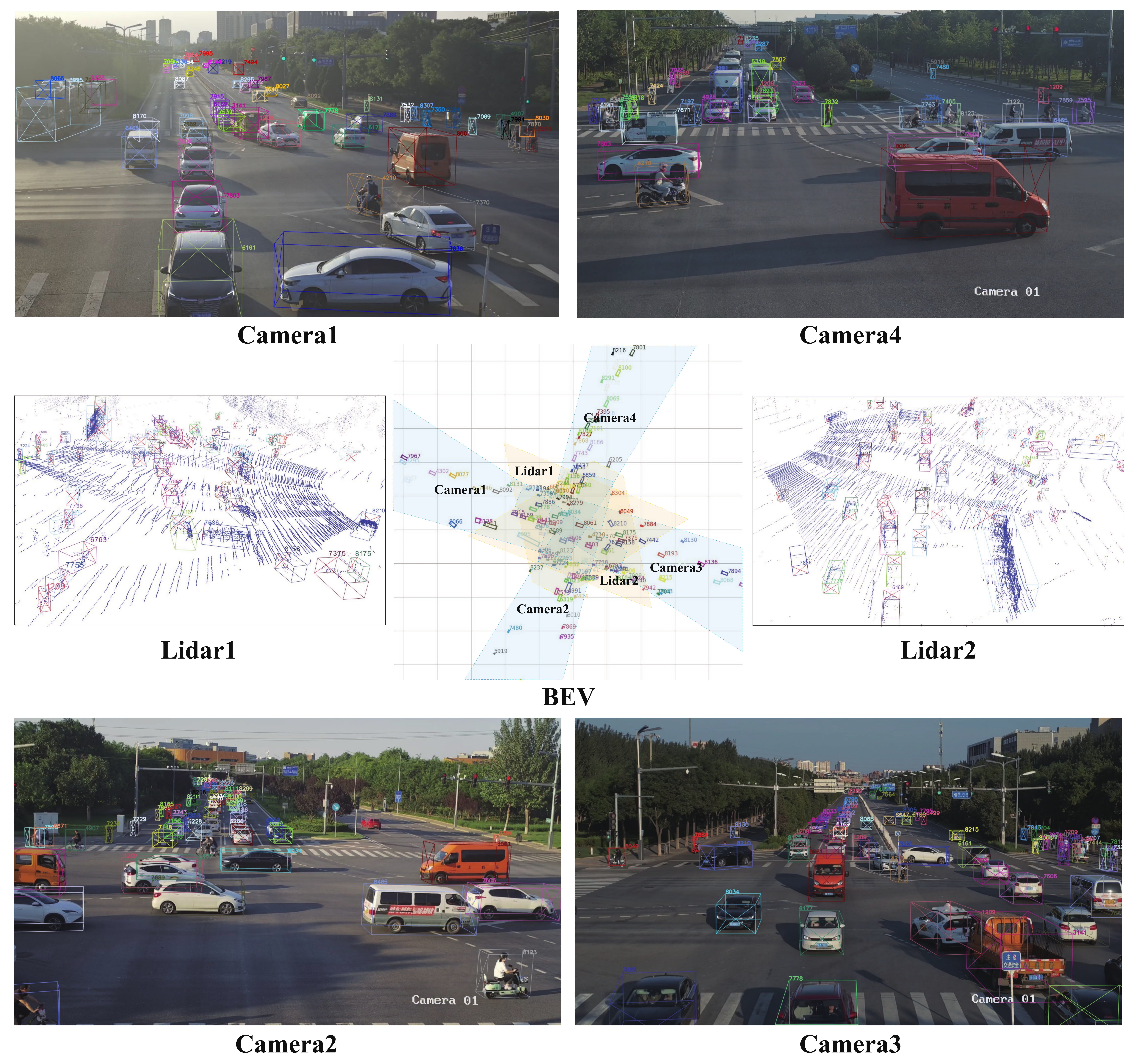}
 \vspace{-0.3cm}
	\caption{An example from HoloVIC dataset: The data and annotated 3D boxes on Camera, Lidar, and BEV, the same targets from different devices are labeled with the same Global ID.
}
	\label{fig1}
\vspace{-0.5cm}
\end{figure}

\begin{table*}[t]\footnotesize
	\begin{center}
\vspace{-0.3cm} 
		\caption{Comparison of popular Datasets in Autonomous Driving and V2X. C: Camera, L: Lidar, F: Fisheye} \label{tab1}
            \setlength{\tabcolsep}{0.3cm}
		\vspace{-0.2cm} 
		\begin{tabular}{c c c c c c c c}
  
			\toprule[1pt]
			Dataset & Source & View  &\makecell{With\\Trajectory} & \makecell{Multi-view\\Overlapping} & \makecell{Sensors Layouts\\ of Vehicle}& \makecell{Sensors Layouts\\ of Infrastructure}& \makecell{Synchronized\\Frames} 
			\\
			\hline
			\hline
			KITTI~\cite{geiger2012we} & real  & vehicle  &\checkmark&  - & 4C+1L&-&14999  \\
   \hline
			nuScenes~\cite{caesar2020nuscenes}& real  & vehicle& \checkmark& - & 6C+1L&-&200k \\
   \hline
			Waymo Open~\cite{sun2020scalability} & real  & vehicle & \checkmark & - & 5C+5L&-&600k   \\
   \hline
			ApolloScape~\cite{huang2018apolloscape} & real  & vehicle  & \checkmark & - & 1L&-&12360 \\
   \hline
               DAIR-V2X-V~\cite{yu2022dair} & real  & vehicle &\ding{55} & - & 2C+2L&- &22325 \\
	   \hline		

			OPV2V~\cite{xu2022opv2v} & simulated  & V2V & \checkmark & -& 4C1L&-&11464  \\
   \hline
			V2VReal~\cite{xu2023v2v4real} & real  & V2V & \checkmark & - & 2C+1L&-&20000  \\
   			
            \hline
            DAIR-V2X-I~\cite{yu2022dair} & real  & infrastructure &\ding{55} & \ding{55}&-& 1C+1L&10084 \\
            \hline
            AICITY22~\cite{Naphade22AIC22} &  real  & infrastructure&\checkmark & \checkmark &-&4C&2132 \\
	       \hline	
            V2X-Sim~\cite{li2022v2x} & simulated  & V2V,VIC& \checkmark & - & 6C+1L&4C+1L&10000  \\
            \hline
            V2XSet~\cite{xu2022v2x} & simulated  & V2V,VIC & \checkmark& \checkmark &-& 4C+1L &11447  \\
            \hline
            			DeepAccident~\cite{wang2023deepaccident} & simulated  &V2V,VIC & \checkmark & \checkmark& 6C+1L&6C+1L& 57000 \\
\hline
   		CARTI~\cite{bai2022pillargrid} & simulated  & VIC  & \checkmark & \checkmark & 1L&1L&11000   \\
     \hline

            DAIR-V2X-C~\cite{yu2022dair} & real  & VIC & \ding{55} & \ding{55} & 2C+2L&1C+1L&38845\\
            \hline
			V2X-seq~\cite{yu2023v2x}& real  & VIC&\checkmark & \ding{55} & 2C+2L&1C+1L & 15000 \\
			\hline
			\hline
			HoloVIC (Ours) & real  & VIC  & \checkmark & \checkmark & 2C+2L&\makecell{4C+2L\\12C+4F+2L} & \makecell{100k}\\
			\bottomrule[1pt]
		\end{tabular}
	\end{center}
\vspace{-0.6cm} 
\end{table*}

\ \ \ Autonomous Driving has experienced notable progression in recent years. In order to further enhance the safety and overall perception, V2X (Vehicle-to-everything) has emerged as a new generation research focus in autonomous driving, which hopefully maximizes the potential of autonomous driving by interaction between Vehicle-to-Vehicle (V2V) and Vehicle-to-Infrastructure (V2I). Currently, Vehicle-Infrastructure Cooperation (VIC) has become a significant research area within V2X. Due to sensors from roadside at a higher viewpoint, the sensors cover wider field compared to the perspective of the vehicle, thus the captured data can provide infomation for the blind spots and farther areas that are beyond the sight of single-vehicle.

V2X has been gradually attracting more attention recently, and some pioneering datasets related to V2X have been released~\cite{li2022v2x,wang2023deepaccident,yu2022dair,yu2023v2x}.  V2X-sim~\cite{li2022v2x} and DeepAccident~\cite{wang2023deepaccident} are generated through simulations using CARLA~\cite{dosovitskiy2017carla} and SUMO~\cite{krajzewicz2012recent}. On the other hand, DAIR-V2X~\cite{yu2022dair} and V2X-seq~\cite{yu2023v2x} are collected from real-world scenarios, but the dataset rely on a single viewpoint by a pair of Camera and Lidar to capture data from different intersections. However, due to the complexity of traffic conditions, the targets captured from Camera are frequently occluded. Therefore, collecting data from a single-viewpoint sensor greatly limits the roadside perception capability.

To further accelerate research in the field of vehicle-infrastructure cooperation, in this paper, we constructed several holographic intersections from diverse perspectives, where the areas captured by multiple sensors overlapping with each other. The intersections consists of 3 different types of sensors (C: Camera, L: Lidar, F: Fisheye). Each intersection is equipped with 6-18 sensors to capture synchronous data. We designed 4 sensor-layouts for different intersections, which includes 4C+2L; 8C+2L; 12C+4F+2L; 4C+2F+2L. Based on these intersections, we build a large-scale multi-viewpoint, multi-sensor dataset and benchmark, named HoloVIC. Meanwhile, autonomous vehicles pass through these intersections and capture data simultaneously with roadside for constructing VIC dataset.

HoloVIC consists of a total of 100k+ frames of synchronized data. Furthermore, the data are obtained from different sensors both on vehicle and road sides. We annotated more than 11.47M 2D\&3D bounding boxes based on 3 types of sensors, and also associate the IDs of the same objects across different devices and consecutive frames in sequence. Then, we formed global trajectories for each individual object from a Bird's-Eye View (BEV) perspective. Based on the annotation of HoloVIC, we generally formulate multiple tasks and benchmark: 1. Monocular 3D Detection (Mono3D); 2. Lidar 3D Detection 3. Multiple Object Tracking (MOT); 4. Multi-sensor Multi-object Tracking (MSMOT); 5. Vehicle-Infrastructure Cooperation Perception (VIC Perception)

The main contributions of our work are as follows:

\begin{itemize}
  \item We constructe several holographic intersections, which adopt 4 different sensor-layouts with Cameras, Fisheyes and Lidars for collecting synchronized data from all sensors in intersections.
  \item We release the first large-scale multi-viewpoint multi-sensor holographic intersection and vehicle-infrastructure cooperation dataset, named HoloVIC.
  \item We annotate 3D bounding boxes on 100k+ synchronized frames based on all the sensors from road-side and vehicle-side, and associate the same targets with unique IDs to form global trajectories.
  \item We formulate five tasks and benchmark to promote the development of research on road-side perception and vehicle-infrastructure cooperative.
\end{itemize}

%% file: sec/2_formatting.tex
\section{Related Work}
\label{sec:formatting}
\quad We summarized open datasets in the field of autonomous driving and V2X, as shown in Tab.\ref{tab1}. Based on the view of the data, datasets are categorized into single vehicle, V2V (Vehicle-to-Vehicle), Infrastructure (Roadside), and Vehicle-Infrastructure Cooperation (VIC). "Multi-view Overlapping" indicates whether the captured areas of roadside sensors overlapping with each other. We have provided information on the number and types of sensors on both the vehicle-side and road-side for each dataset, as well as the number of captured synchronized frames in the dataset.


\begin{figure*}
	\centering
	\includegraphics[width=17.5cm]{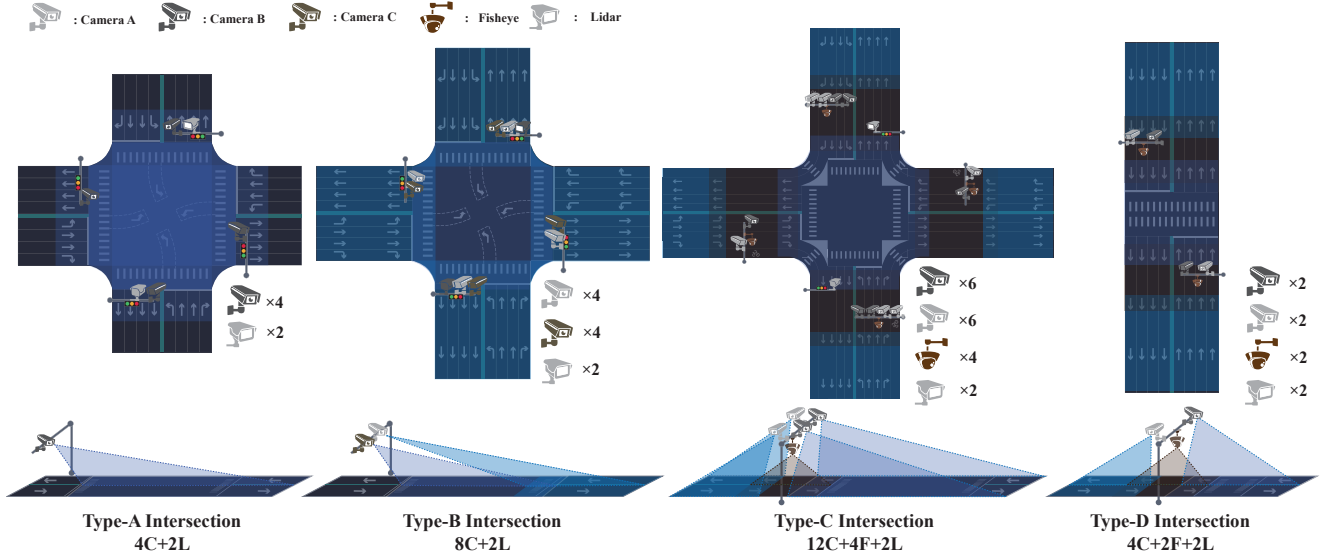}
 \vspace{-0.5cm}
	\caption{The configuration of holographic intersections: The figure illustrates three different sensors (C: Camera, L: Lidar, F: Fisheye) and four various sensor-layouts (4C+2L, 8C+2L, 12C+4F+2L, 4C+2F+2L) in holographic intersections}
	\label{fig2}
\vspace{-0.4cm}
\end{figure*}

\textbf{Vehicle-Side Datasets}~\cite{geiger2012we,caesar2020nuscenes,sun2020scalability,huang2018apolloscape} primarily serve perception algorithms for autonomous driving. Common single-vehicle datasets include KITTI~\cite{geiger2012we}, a pioneering dataset that includes synchronized video frames obtained from stereo Cameras and Lidar, totaling 14,999 frames. It is used for pointcloud detection and tracking. The ApolloScape~\cite{huang2018apolloscape} includes additional pixel-wise masks, which is utilized for vehicle-side segmentation. The Waymo~\cite{caesar2020nuscenes} contains a large amount of Camera and Lidar data , including pointclouds, images, and vehicle localization, it is primarily used for perception, prediction, and planning. And nuScenes~\cite{caesar2020nuscenes} provides sensor data from a variety of sources, including Lidar, Radar, and Camera, with a significant number of calibrated frames. 


\textbf{Simulated V2X Datasets}~\cite{xu2022opv2v,li2022v2x,xu2022v2x,wang2023deepaccident} are generated by simulators. At the beginning of V2X, researchers utilized simulators such as CARLA~\cite{dosovitskiy2017carla}, SUMO~\cite{krajzewicz2012recent}, and OpenCDA~\cite{xu2021opencda} to generate V2V and VIC data. The simulators are able to generate large-scale V2X data easily. OPV2V~\cite{xu2022opv2v} is the first large-scale open simulated dataset for Vehicle-to-Vehicle perception. V2X-sim~\cite{li2022v2x} and V2X-set~\cite{xu2022v2x} employs simulators to obtain traffic flows and collect sensor stream. DeepAccident~\cite{wang2023deepaccident} utilizes CARLA to generate approximately 7 times more than nuScenes, and formulated end-to-end motion and accident prediction task.

\textbf{Real-world Roadside and VIC Datasets}~\cite{yu2022dair,yu2023v2x,chavdarova2018wildtrack,Naphade22AIC22} are difficult to be collected and annotated due to the dependency on infrastructure development. However, researchers are not satisfied with simulation datasets. In recent years, real-world datasets have been released. DAIR-V2X~\cite{yu2022dair} is the first large-scale real-world VIC dataset, and V2X-Seq~\cite{yu2023v2x} is the first VIC sequential perception dataset. Additionally, multi-sensor with overlapping Datasets in road-side have been released such as WildTrack~\cite{chavdarova2018wildtrack} and AICITY2022~\cite{Naphade22AIC22}. These datasets involve synchronized data streams from 4-7 Cameras and provide annotations for the same objects across multiple Cameras. Due to the difficulty of annotation, these datasets only provide annotations for a few thousand synchronized frames.

%% file: sec/3_finalcopy.tex
\section{HoloVIC Open Dataset}

\subsection{Multi-Sensor Layouts}
\quad Due to the varying width of road, number of lanes, and shapes of intersection, we adopt four distinct sensor layouts to ensure optimal coverage of the intersection areas, as shown in Fig.\ref{fig2}.

\textbf{Type-A Intersection (4C+2L)} utilizes four Cameras mounted on signal poles in all four directions to perceive the central area of the intersection. Additionally, two Lidars are deployed in opposing directions in signal poles to form a layout of 4C+2L.

\textbf{Type-B Intersections (8C+2L)} encompass the central area of the intersection and the area beyond the stop line to increase the coverage of perception. We have installed a set of short-focus and telephoto Cameras on signal poles in all four directions to capture images of both the inside and outside of intersection respectively. Furthermore, two Lidar sensors are mounted in opposing directions, resulting in a sensor layout of 8C+2L.

\begin{figure*}
	\centering

	\includegraphics[width=17.7cm]{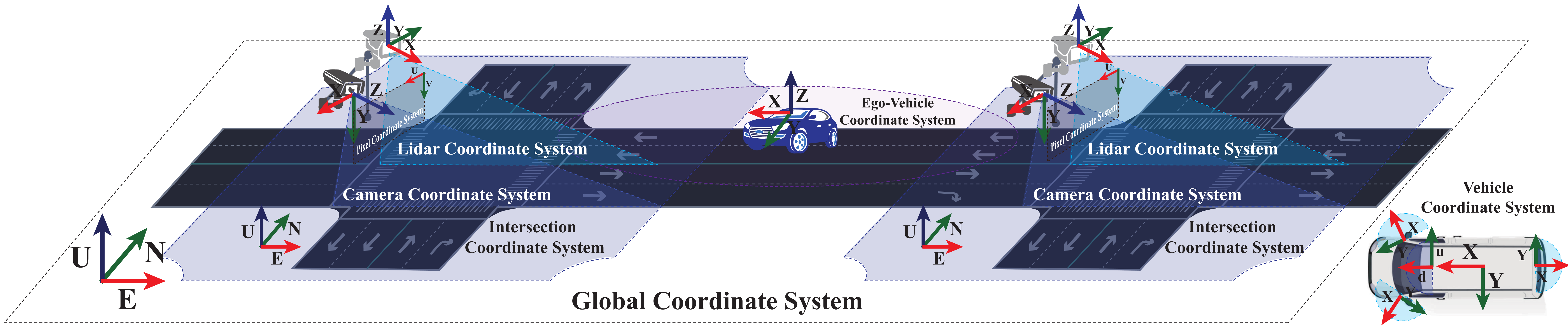}
  \vspace{-0.5cm}
	\caption{The coordinate systems in the HoloVIC dataset, involving all sensors on both vehicle and road sides.}
	\label{fig3}
\vspace{-0.4cm}
\end{figure*}

\textbf{Type-C Intersections (12C+4F+2L)} have more lanes. To ensure better coverage of both the inside and outside areas of the intersection, we deploy 2 Cameras on both sides of the monitoring poles for four directions.  In addition, to account for blind spots directly beneath the monitoring poles that are not captured by the Cameras, we install a Fisheye on each pole to effectively cover these blind spots. Similarly, a set of two Lidars are deployed in opposing directions, and the layout of type-C is 12C+4F+2L.

\textbf{Type-D intersections (4C+2F+2L)} only allows pedestrians cross laterally, while vehicles are restricted to going straight or making U-turns. Therefore, similar to the installation plan for Type C intersections, we install two Cameras and Fisheye in each of the two directions. And we deploy two Lidars to compose 4C+2F+2L.

\subsection{Coordinate Systems}
\quad We introduce all of the coordinate systems involved in dataset as shown in Fig.\ref{fig3}. All of the coordinate systems follow the right-hand rule. The definition, calibration and transformation of coordinate systems are as follows:

\textbf{Global Coordinate:} To align the coordinates of intersections and vehicles, we select a point within the entire range of intersections as the origin of the global coordinate$(\omega_{x_0},\omega_{y_0},\omega_{z_0})$. The global coordinate is constructed based on the East-North-Up (ENU) coordinate system. Any position $(\omega_x,\omega_y,\omega_z)$ within the scene is calculated the relative actual distance to the origin in east-west and north-south as the x-axis and y-axis, respectively. The distance between two points is based on the WGS84, which is collected by Real-Time Kinematic (RTK). Both road and vehicle are aligned by converting their respective coordinate systems to the global coordinate system.

\textbf{Intersection Coordinate} also utilizes ENU coordinate system $(\sigma_x,\sigma_y,\sigma_z)$. We select a point as the origin of the intersection coordinate system for each intersection. The Intersection Coordinate System is the most crucial coordinate system for roadside perception, which is used for aligning all the sensor coordinate systems from the intersection.

\textbf{Ego-Vehicle Coordinate} $(\delta_x,\delta_y,\delta_z)$ is defined with the vehicle center as the origin. The forward, left and upward direction represent the positive X,Y and Z,respectively. All sensors on the vehicle will eventually be converged into the vehicle coordinate system, which further utilized for vehicle-infrastructure cooperative perception.

\textbf{Lidar Coordinate:} is a 3D coordinate system that includes the $x$, $y$, and $z$ dimensions to represent the spatial position of pointclouds $(x,y,z)$. The transformation between the Lidar and Intersection/Ego-Vechile coordinate can be achieved by calibrating multiple points between different coordinate systems and solving for the rotation $R_L\in\mathbb{R}^{3\times3}$ and translation $T_L\in\mathbb{R}^{3\times1}$ by Kabsch~\cite{kabsch1976solution}, projection equations in homogeneous coordinates is defined as:

\begin{eqnarray}
\begin{pmatrix}
x\\
y\\
z \\
1
\end{pmatrix}=
\begin{bmatrix}
{R}^{3\times3}_L&{T}^{3\times1}_L\\ 
0&1
\end{bmatrix}
\begin{pmatrix}
\sigma_x\\
\sigma_y\\
\sigma_z \\
1
\end{pmatrix}
\end{eqnarray}

\textbf{Camera Coordinate:} is a 3D coordinate system, where the $z$-axis represents the depth in the Camera Coordinate System $(c_x,c_y,c_z)$. The transformation from the intersection coordinate system to the Camera Coordinate System can be defined as:
\begin{eqnarray}
\begin{pmatrix}
c_x\\
c_y\\
c_z \\
1
\end{pmatrix}=
S^{4\times4}
\begin{bmatrix}
{R}^{3\times3}_C&{T}^{3\times1}_C\\ 
0&1
\end{bmatrix}
\begin{pmatrix}
\sigma_x\\
\sigma_y\\
\sigma_z \\
1
\end{pmatrix}
\end{eqnarray}

where $S\in\mathbb{R}^{4\times4}$ is used for mapping between Intersection-Camera Coordinates axes. Rotation ${R}^{3\times3}_C$ and Translation ${T}^{3\times1}_C$ are calculated by solving PnP (Perspective-n-Point).

\textbf{Pixel Coordinate} is obtained by projecting the Camera coordinate onto the imaging plane, which is transformed into 2D coordinate system $(u,v)$. The transformation from Camera coordinate to pixel coordinate is formulated as:

\begin{eqnarray}
Z_c
\begin{pmatrix}
u\\
v\\
1
\end{pmatrix}=
K^{3\times3}
\begin{pmatrix}
c_x\\
c_y\\
c_z 
\end{pmatrix}
,\ 
K^{3\times3}=
\begin{bmatrix}
f_x&-1&u_0\\
0&f_y&v_0\\
0&0&1\\
\end{bmatrix}
\end{eqnarray}
$K\in\mathbb{R}^{3\times3}$ indicates the intrinsic matrix of Camera, $f_x,f_y$ donote the focal of the Camera in $x$-axis, $y$-axis. $u_0,v_0$ represent the center of image. We calibrated each Camera before capturing the data by Chessboard Calibration.

\subsection{Ground Truth Labels}
\quad In the HoloVIC dataset, we provide high-quality annotation, the ground truth include: 

1. We annotated 3D bounding boxes $[x,y,z,l,w,h,\theta]$ and category in both the Camera coordinate for images and in the Lidar coordinate for pointclouds, where $x,y,z$ represent the center of 3D box in 3D coordinate, while $l,w,h$ correspond to length, width and height, and $\theta$ is the orientation (yaw) of 3D box. The category $\eta$ include "Vehicle", "Cyclist" and "Pedestrian".

2. We assigned Track ID to the same target across the temporal sequence of each sensor. For the same target within a sequence, it has unique tracking ID, $\tau$. The bounding boxes with the same $\tau$ are linked together to form a complete trajectory.

3. We associated global ID to the same object across different sensors at the same timestamp. For the same target within a intersection, it only has unique global ID $\rho$. Additionally, we also generate the 3D box of Target $\rho$ in intersection coordinate $\rho:[\sigma_x,\sigma_y,\sigma_z,l,w,h,\theta,\eta]$ based on all of the boxes with global ID of $\rho$. 

4. We matched the the same objects between from the vehicle-side $\nu$ and road-side $\rho$ as the unique global ID.

%% file: sec/4_Tasks.tex
\section{Tasks \& Metrics}
\begin{table*}[]\footnotesize\centering
\begin{center}
\caption{The details of HoloVIC dataset, including the proportion allocation between Infrastructure and VIC; The proportion of different scenes; The distribution of training, testing and validation sets;  The count of synchronized frames and annotated 3D boxes for each scene.} \label{tab3}
\vspace{-0.2cm}
\setlength{\tabcolsep}{0.34cm}
\begin{tabular}{c|c|c|cc|ccc|c|c|c}
\toprule[1pt]
\multirow{2}{*}{View} & \multirow{2}{*}{Scene} & \multirow{2}{*}{Layout} & \multicolumn{2}{c|}{Distribution} & \multicolumn{3}{c|}{Ratio of Dataset} & \multirow{2}{*}{\begin{tabular}[c]{@{}c@{}}Num of \\ Frames\end{tabular}} & \multirow{2}{*}{\begin{tabular}[c]{@{}c@{}}3D Boxes\\ (Global)\end{tabular}} & \multirow{2}{*}{\begin{tabular}[c]{@{}c@{}}3D Boxes\\ (Local)\end{tabular}} \\ \cline{4-8}
 &  &  & \multicolumn{1}{c|}{View} & Scene & \multicolumn{1}{c|}{Train} & \multicolumn{1}{c|}{Test} & Valid &  &  &  \\ \hline
\multirow{5}{*}{\begin{tabular}[c]{@{}c@{}}Infrastructure\end{tabular}} & Int-1 & 4C+2L & \multicolumn{1}{c|}{\multirow{5}{*}{70\%}} & 30\% & \multicolumn{1}{c|}{\multirow{4}{*}{50\%}} & \multicolumn{1}{c|}{\multirow{4}{*}{40\%}} & \multirow{4}{*}{10\%} & 21k & 600k & 2M \\ \cline{2-3} \cline{5-5} \cline{9-11} 
 & Int-2 & 8C+2L & \multicolumn{1}{c|}{} & 30\% & \multicolumn{1}{c|}{} & \multicolumn{1}{c|}{} &  & 21k & 480k & 3.6M \\ \cline{2-3} \cline{5-5} \cline{9-11} 
 & Int-3 & 12C+4F+2L & \multicolumn{1}{c|}{} & 10\% & \multicolumn{1}{c|}{} & \multicolumn{1}{c|}{} &  & 7k & 85k & 1.2M \\ \cline{2-3} \cline{5-5} \cline{9-11} 
 & Int-4 & 4C+2F+2L & \multicolumn{1}{c|}{} & 20\% & \multicolumn{1}{c|}{} & \multicolumn{1}{c|}{} &  & 14k & 120k & 420k \\ \cline{2-3} \cline{5-11} 
 & Int-5 & 12C+4F+2L & \multicolumn{1}{c|}{} & 10\% & \multicolumn{1}{c|}{0\%} & \multicolumn{1}{c|}{100\%} & 0\% & 7k & 87k & 1.6M \\ \hline
\multirow{5}{*}{\begin{tabular}[c]{@{}c@{}}VIC\end{tabular}} & VIC-1 & 4C+2L & \multicolumn{1}{c|}{\multirow{5}{*}{30\%}} & 30\% & \multicolumn{1}{c|}{\multirow{4}{*}{50\%}} & \multicolumn{1}{c|}{\multirow{4}{*}{40\%}} & \multirow{4}{*}{10\%} & 18k & 210k & 650k \\ \cline{2-3} \cline{5-5} \cline{9-11} 
 & VIC-2 & 8C+2L & \multicolumn{1}{c|}{} & 30\% & \multicolumn{1}{c|}{} & \multicolumn{1}{c|}{} &  & 18k & 100k & 580k \\ \cline{2-3} \cline{5-5} \cline{9-11} 
 & VIC-3 & 12C+4F+2L & \multicolumn{1}{c|}{} & 10\% & \multicolumn{1}{c|}{} & \multicolumn{1}{c|}{} &  & 3k & 40k & 580k \\ \cline{2-3} \cline{5-5} \cline{9-11} 
 & VIC-4 & 4C+2F+2L & \multicolumn{1}{c|}{} & 20\% & \multicolumn{1}{c|}{} & \multicolumn{1}{c|}{} &  & 6k & 65k & 205k \\ \cline{2-3} \cline{5-11} 
 & VIC-5 & 12C+4F+2L & \multicolumn{1}{c|}{} & 10\% & \multicolumn{1}{c|}{0\%} & \multicolumn{1}{c|}{100\%} & 0\% & 3k & 45k & 640k \\ \hline \hline
Total & - & - & \multicolumn{2}{c|}{100\%} & \multicolumn{1}{c|}{45\%} & \multicolumn{1}{c|}{46\%} & 9\% & 100k & 1.8M & 11.47M \\ 
\bottomrule[1pt]
\end{tabular}
\end{center}
\vspace{-0.6cm}
\end{table*}

\quad Based on HoloVIC data and annotation, we formulated the tasks based on Single-Sensor, Multi-Sensor and VIC Perception, which are introduced as following.

\subsection{Single-Sensor Perception}
\subsubsection{3D Detection}

\quad The 3D Detection task of the HoloVIC includes Monocular 3D (Mono3D) and Lidar 3D Detection. Given an image frame or pointcloud, the detection model is used to obtain the position, shape, orientation, and category of targets in the form of 3D bounding boxes $[x,y,z,l,w,h,\theta,\eta]$. We refer to the metrics from Rope3D~\cite{ye2022rope3d}, KITTI~\cite{geiger2012we}, and nuScenes~\cite{geiger2012we} and adopt mAP (mean Average Precision) and mAOS (mean Average Orientation Similarity) to evaluate detectors for both the tasks.

Mean Average Precision (mAP) is used to evaluate the accuracy of object detection. The mAP score is influenced by the position, size, orientation, classification, and confidence of the predicted bounding boxes, which is defined as:

\begin{eqnarray}
mAP=\frac{1}{\mathbb{C}}\sum_{\eta\in\mathbb{C}}{AP}_{\eta}
\label{eq4}
\end{eqnarray} 
where $\mathbb{C}$ is the set of category, $AP$ indicates Average Precision~\cite{everingham2010pascal}. For 2D Detection task, we utilize 3D intersection over union (3D IOU) to match the prediction and ground truth, which is formulated as:

\begin{eqnarray}
AP|_n=\frac{1}{|Rc|}\sum_{r\in Rc} \max_{\widetilde{r}:\widetilde{r}>r}Pr(\widetilde{r})
\label{eq5}
\end{eqnarray} 
where $Pr(\widetilde{r})$ is the precision at a certain recall threshold $r \in \{\frac{1}{n},\frac{2}{n},...,1\}$, $|Rc|=n+1$ indicates the number of elements in $Rc$, we set $n$ to 40.

Mean Average Orientation Similarity is utilized to measure the precision of yaw angle, which is defined as:
\begin{eqnarray}
mAOS=\frac{1}{\mathbb{C}}\sum_{\eta\in\mathbb{C}}{AOS}_{\eta}
\label{eq6}
\end{eqnarray} 
where $AOS$ (Average Orientation Similarity)~\cite{sun2020scalability}, for 3D Detection task, we only consider the angle around the z-axis rotation (yaw), which is formulated as:
\begin{eqnarray}
AOS|_n=\frac{1}{|Rc|}\sum_{r\in Rc} \max_{\widetilde{r}:\widetilde{r}>r}Ori(\widetilde{r})
\label{eq7}
\end{eqnarray} 
\begin{eqnarray}
Ori(r)=\frac{1}{|D(r)|}\sum_{i\in D(r)}\frac{1+cos\Delta_\theta^{(i)}}{2}\delta_i
\label{eq8}
\end{eqnarray} 
where $D$ is set of true positive samples, $\Delta_\theta^{(i)}$ is the angle difference of sample $i$. To penalize multiple boxes matching to the same ground truth, we set $\delta_i=1$ for box $i$ if it has already been matched to a ground truth, otherwise $\delta_i=0$.

\subsubsection{Tracking}

\quad The Tracking task of the HoloVIC consists of 2D Tracking on videos and 3D Tracking both on video and pointcloud sequences. Given the sequence and the detection result as inputs, tracking model aims to associate the bounding boxes for same target across the sequence and assign Track ID $\tau$ for each target. We refer to the MOT metrics~\cite{bernardin2008evaluating,ristani2016performance} and adopt MOTA (MOT Accuracy), IDF1 (ID F1 Score) to measure Tracking task.

The definition of MOTA metric is as follows:

\begin{eqnarray}
MOTA=1-\frac{|FP|+|FN|+|IDSw|}{|gtDet|}
\label{eq9}
\end{eqnarray} 
where FP (False Positive) and FN (False Negative) indicate the wrongly detected and missed from detection. IDSw (ID switch) represents the tracker incorrectly assign different IDs for same target or swap the IDs of two objects. We also evaluate MT (Mostly Tracked), and ML (Mostly Lost) as additional reference metrics from CLEAR-MOT~\cite{bernardin2008evaluating}.

IDF1 calculates one-to-one mapping the Trajectories between prediction and ground truth, which is defined as:
\begin{eqnarray}
IDF1=\frac{2|IDTP|}{2|IDTP|+|IDFP|+|IDFN|}
\label{eq10}
\end{eqnarray} 
where IDTP (identity true positives) are matches on overlapping part of trajectories that are matched. IDFN (identity false negatives) and IDFP (identity false positives) are calculated from both non-overlapping of matched trajectories, and the remaining trajectories that are not matched. We also evaluate IDP (ID-Precision) and IDR (ID-Recall) as reference metrics from Identity Metrics~\cite{ristani2016performance}.

\subsection{Multi-Sensor Perception}
\quad The Multi-Sensor Perception task is primarily used to analyze the overall situation at intersections. It involves multiple sensors capturing data simultaneously in intersections. The perception task is divided into detection and tracking, which utilize data and the spatial-temporal relationships between devices to output perception results in the intersection coordinate system, which are presented in the form of 3D boxes $[\sigma_x,\sigma_y,\sigma_z,l,w,h,\theta,\eta]$. The ground truth 3D boxes from each sensor are merged in the intersection coordinate system to create ground truth for the multi-sensor perception task, which is defined as:
\begin{eqnarray}
GT_i=\{GT_s,s\in S\}
\end{eqnarray}
where $GT_s$ is the ground truth in $s$-th Sensor, $S$ is the set of Sensors in Intersection. We refer to 3D Detection metrics~\cite{geiger2012we} and adopt mAP and mAOS as the evaluation metrics for multi-sensor 3D detection. In addition, we utilized metrics of Multi-target Multi-camera Tracking (MTMCT)~\cite{ristani2016performance} and selected MOTA and IDF1 as the metrics for multi-sensor 3D Tracking. All the definitions of metrics are consistent with the relevant descriptions in Eq.\ref{eq4}-\ref{eq10}.

\subsection{VIC Perception}
\quad The VIC perception task focuses on the cooperative perception between vehicles and infrastructure. Given synchronized data from both on vehicle and road sides, VIC is used to evaluate the capability fusing information and assess the benefits brought to vehicles by roadside perception.

We firstly align the position of the 3D boxes from the intersection coordinate to the ego-vehicle coordinate. The ground truth 3D boxes are merged from both vehicle-side and road-side at the same timestamp $t$:

\begin{eqnarray}
GT_{vic}^t=GT_{v}^t\cup GT_{i|v}^t
\end{eqnarray}
\begin{eqnarray}
GT_{i|v}^t=\{\rho_{i\rightarrow v}\in GT_{i}^t, ||\rho_{i\rightarrow v}[\sigma]-\nu[\sigma]||<\varepsilon\}
\end{eqnarray}
where $GT_{i}^t$ is the ground truth from roadside at $t$, $\rho_{i\rightarrow v}$ indicates the position of target from roadside after transformed to the ego vehicle coordinate. We defined a range around the position of ego vehicle $\nu[\sigma]$, any roadside 3D box that exceeds the distance $\varepsilon$ is discarded. We refer to the metrics from DAIR-V2X~\cite{yu2022dair,yu2023v2x} and adopt mAP, mAOS for Detection, MOTA and IDF1 for Tracking. We evaluate results separately with same ground truth $GT_{vic}^t$: 1. ONLY vehicle-side data as input 2. Both vehicle-side and roadside as input. This approach allows for a better comparison to quantify the benefits brought to ego vehicle perception by incorporating roadside data.

%% file: sec/5_experiments.tex
\section{Experiments}
\begin{table*}[]\footnotesize
\begin{center}
\vspace{-0.4cm}
\caption{Mono3D and Lidar 3D Detection results on validation sets} \label{tab4}
\vspace{-0.3cm}
\setlength{\tabcolsep}{0.32cm}
\begin{tabular}{c|c|ccc|ccc|ccc}
\toprule[1pt]
\multirow{2}{*}{Task} & \multirow{2}{*}{Method} & \multicolumn{3}{c|}{Vehicle@0.2/0.7} & \multicolumn{3}{c|}{Cyclist@0.2/0.5} & \multicolumn{3}{c}{Pedestrian@0.2/0.5} \\ \cline{3-11} 
 &  & \multicolumn{1}{c|}{3d AP} & \multicolumn{1}{c|}{bev AP} & AOS & \multicolumn{1}{c|}{3d AP} & \multicolumn{1}{c|}{bev AP} & AOS & \multicolumn{1}{c|}{3d AP} & \multicolumn{1}{c|}{bev AP} & AOS \\ \hline \hline
\multirow{2}{*}{\begin{tabular}[c]{@{}c@{}}Mono3D\\ Detection\end{tabular}} & FCOS3D~\cite{wang2021fcos3d} & \multicolumn{1}{c|}{47.21} & \multicolumn{1}{c|}{37.17} & 68.05 & \multicolumn{1}{c|}{11.68} & \multicolumn{1}{c|}{11.60} & 47.96 & \multicolumn{1}{c|}{11.31} & \multicolumn{1}{c|}{6.99} & 53.79 \\ \cline{2-11} 
 & PGD~\cite{wang2022probabilistic} & \multicolumn{1}{c|}{52.54} & \multicolumn{1}{c|}{43.21} & \textbf{74.23} & \multicolumn{1}{c|}{17.31} & \multicolumn{1}{c|}{19.15} & 53.18 & \multicolumn{1}{c|}{8.67} & \multicolumn{1}{c|}{4.78} & \textbf{58.35} \\ \hline
\multirow{4}{*}{\begin{tabular}[c]{@{}c@{}}Lidar 3D\\ Detection\end{tabular}} & SECOND~\cite{yan2018second} & \multicolumn{1}{c|}{70.54} & \multicolumn{1}{c|}{81.35} & 60.84 & \multicolumn{1}{c|}{77.35} & \multicolumn{1}{c|}{80.86} & 57.80 & \multicolumn{1}{c|}{26.25} & \multicolumn{1}{c|}{29.81} & 29.90 \\ \cline{2-11} 
 & Pointpillars~\cite{lang2019pointpillars} & \multicolumn{1}{c|}{70.17} & \multicolumn{1}{c|}{81.69} & 60.61 & \multicolumn{1}{c|}{73.04} & \multicolumn{1}{c|}{78.84} & 57.8 & \multicolumn{1}{c|}{27.43} & \multicolumn{1}{c|}{31.31} & 24.76 \\ \cline{2-11} 
 & PVRCNN~\cite{shi2020pv} & \multicolumn{1}{c|}{\textbf{72.78}} & \multicolumn{1}{c|}{82.16} & 60.50 & \multicolumn{1}{c|}{77.06} & \multicolumn{1}{c|}{\textbf{82.22}} & 58.01 & \multicolumn{1}{c|}{21.91} & \multicolumn{1}{c|}{24.47} & 22.54 \\ \cline{2-11} 
 & Transfusion-L~\cite{bai2022transfusion} & \multicolumn{1}{c|}{69.79} & \multicolumn{1}{c|}{\textbf{82.86}} & 60.49 & \multicolumn{1}{c|}{\textbf{77.51}} & \multicolumn{1}{c|}{81.64} & \textbf{58.32} & \multicolumn{1}{c|}{\textbf{45.51}} & \multicolumn{1}{c|}{\textbf{52.57}} & 44.54 \\ 
 \bottomrule[1pt]
\end{tabular}
\end{center}
\vspace{-0.4cm}
\end{table*}

\begin{table*}[]\footnotesize
\begin{center}
\vspace{-0.2cm}
\caption{The performance of 2D/3D MOT on validation sets} \label{tab5}
\vspace{-0.3cm}
\setlength{\tabcolsep}{0.4cm}
\begin{tabular}{c|c|c|cccccc}
\toprule[1pt]
Task & Detector & Tracker & IDF1$\uparrow$ & IDP$\uparrow$ & IDR$\uparrow$ & MOTA$\uparrow$ & MT$\uparrow$ & ML$\downarrow$ \\ \hline\hline
\multirow{3}{*}{2D MOT} & \multirow{3}{*}{YOLOv8~\cite{reis2023real}} & DeepSort~\cite{wojke2017simple} &53.72  &67.41  &45.53  & 41.72 & 26.38 & 34.74 \\ \cline{3-9} 
 &  & Tracktor~\cite{bergmann2019tracking} & 59.34 & 71.58 & 53.69 & 57.70 & 39.85 & 24.21 \\ \cline{3-9} 
 &  & FairMOT~\cite{zhang2021fairmot} & \textbf{68.17} & \textbf{72.35} & \textbf{62.57} & \textbf{63.35} & \textbf{47.98} & \textbf{17.45} \\ \hline
\begin{tabular}[c]{@{}c@{}}Mono3D MOT\end{tabular} & FCOS3D~\cite{wang2021fcos3d} & AB3DMOT~\cite{2020AB3DMOT} & 42.85 & 57.36 & 32.17 & 39.34 & 21.81 & 27.37 \\ \hline
\begin{tabular}[c]{@{}c@{}}Lidar 3D MOT\end{tabular} & Pointpillars~\cite{lang2019pointpillars} & AB3DMOT~\cite{2020AB3DMOT} & 57.62 & 69.59 & 43.86 & 51.45 & 24.70 & 30.71 \\ 
\bottomrule[1pt]
\end{tabular}
\end{center}
\vspace{-0.5cm}
\end{table*}

\begin{table}\footnotesize\centering
	\begin{center}
       \vspace{-0.2cm}
   \caption{Sensor Specifications in HoloVIC} \label{tab2}
      \vspace{-0.2cm}
        \setlength{\tabcolsep}{0.35cm}
		\begin{tabular}{c|c|c}
			\toprule[1pt]
			Scene& Sensor  & Details 
			\\
			\hline
            \hline
                 			\multirow{5}{*}{\makecell{ \\ \\ \\ \\Infrastructure}}
			&Camera A&\makecell{RGB, 25hz, 1920$\times$ 1080\\ FOV:$[{31.6}^{\circ},{17.0}^{\circ}]$}\\
			\cline{2-3}
   			&Camera B&\makecell{RGB, 25hz, 1920$\times$ 1080\\ FOV:$[{47.7}^{\circ},{25.2}^{\circ}]$}\\
			\cline{2-3}
      			&Camera C&\makecell{RGB, 25hz, 1920$\times$ 1080\\ FOV:$[{111.78}^{\circ},{63.16}^{\circ}]$}\\
			\cline{2-3}
      			&Fisheye &\makecell{RGB, 25hz, 2048$\times$ 2048\\ FOV:$[{180.0}^{\circ},{180.0}^{\circ}]$}\\
			\cline{2-3}
			&Lidar&\makecell{150 beams, 10hz, 1.6M pps\\ FOV:$[{100.0}^{\circ},{40.0}^{\circ}]$}\\
   			\hline
                 			\multirow{4}{*}{\makecell{ \\ \\ \\Vehicle}}
			&Camera A&\makecell{RGB, 25hz, 1920$\times$ 1080\\ FOV:$[{30.0}^{\circ},{17.0}^{\circ}]$}\\
			\cline{2-3}
   			&Camera B&\makecell{RGB, 25hz, 1920$\times$ 1080\\ FOV:$[{120.0}^{\circ},{17.0}^{\circ}]$}\\
            \cline{2-3}
      			&Lidar A&\makecell{40 beams, 10hz, 720k pps\\ FOV:$[{286.48}^{\circ}$,${-25}^{\circ}$ to ${15}^{\circ}]$}\\
			\cline{2-3}
   			&Lidar B&\makecell{64 beams, 10hz, 384k pps\\ FOV:$[{360.0}^{\circ}$,${-25}^{\circ}$ to ${15}^{\circ}]$}\\
            \bottomrule[1pt]
			
		\end{tabular}

	\end{center}
 \vspace{-0.6cm}
\end{table}

\begin{table*}[]\footnotesize
\begin{center}
\vspace{-0.2cm}
\caption{Detection and tracking accuracy of only-vehicle, VIC with "1C+1L" and "4C+2L" on Int-1 and VIC-1 validation sets.} \label{tab6}
\vspace{-0.2cm}
\setlength{\tabcolsep}{0.48cm}
\begin{tabular}{c|cccc|cccc}

\hline
\multirow{3}{*}{View} & \multicolumn{4}{c|}{Detection} & \multicolumn{4}{c}{Tracking} \\ \cline{2-9} 
 & \multicolumn{1}{c|}{\multirow{2}{*}{Metric}} & \multicolumn{3}{c|}{Range(m)} & \multicolumn{1}{c|}{\multirow{2}{*}{IDF1$\uparrow$}} & \multicolumn{1}{c|}{\multirow{2}{*}{MOTA$\uparrow$}} & \multicolumn{1}{c|}{\multirow{2}{*}{MT$\uparrow$}} & \multirow{2}{*}{ML$\downarrow$} \\ \cline{3-5}
 & \multicolumn{1}{c|}{} & \multicolumn{1}{c|}{{[}0,30{]}} & \multicolumn{1}{c|}{{[}30,50{]}} & {[}50,70{]} & \multicolumn{1}{c|}{} & \multicolumn{1}{c|}{} & \multicolumn{1}{c|}{} &  \\ \hline
\multirow{2}{*}{\begin{tabular}[c]{@{}c@{}}Vehicle\\ (V-only)\end{tabular}} & \multicolumn{1}{c|}{bev AP} & \multicolumn{1}{c|}{97.85} & \multicolumn{1}{c|}{90.97} & 46.95 & \multicolumn{1}{c|}{\multirow{2}{*}{86.57}} & \multicolumn{1}{c|}{\multirow{2}{*}{75.64}} & \multicolumn{1}{c|}{\multirow{2}{*}{50.01}} & \multirow{2}{*}{15.38} \\ \cline{2-5}
 & \multicolumn{1}{c|}{AOS} & \multicolumn{1}{c|}{98.16} & \multicolumn{1}{c|}{94.01} & 49.67 & \multicolumn{1}{c|}{} & \multicolumn{1}{c|}{} & \multicolumn{1}{c|}{} &  \\ \hline
\multirow{2}{*}{\begin{tabular}[c]{@{}c@{}}VIC\\ (V+I: 1C+1L)\end{tabular}} & \multicolumn{1}{c|}{bev AP} & \multicolumn{1}{c|}{\textbf{98.09}} & \multicolumn{1}{c|}{91.51} & 59.82 & \multicolumn{1}{c|}{\multirow{2}{*}{90.02}} & \multicolumn{1}{c|}{\multirow{2}{*}{81.19}} & \multicolumn{1}{c|}{\multirow{2}{*}{61.54}} & \multirow{2}{*}{7.69} \\ \cline{2-5}
 & \multicolumn{1}{c|}{AOS} & \multicolumn{1}{c|}{\textbf{98.42}} & \multicolumn{1}{c|}{94.21} & 63.00 & \multicolumn{1}{c|}{} & \multicolumn{1}{c|}{} & \multicolumn{1}{c|}{} &  \\ \hline
\multirow{2}{*}{\begin{tabular}[c]{@{}c@{}}VIC\\ (V+I 4C+2L)\end{tabular}} & \multicolumn{1}{c|}{bev AP} & \multicolumn{1}{c|}{\textbf{98.09}} & \multicolumn{1}{c|}{\textbf{92.06}} & \textbf{85.75} & \multicolumn{1}{c|}{\multirow{2}{*}{\textbf{95.47}}} & \multicolumn{1}{c|}{\multirow{2}{*}{\textbf{90.19}}} & \multicolumn{1}{c|}{\multirow{2}{*}{\textbf{92.31}}} & \multirow{2}{*}{\textbf{3.85}} \\ \cline{2-5}
 & \multicolumn{1}{c|}{AOS} & \multicolumn{1}{c|}{\textbf{98.42}} & \multicolumn{1}{c|}{\textbf{94.75}} & \textbf{90.43} & \multicolumn{1}{c|}{} & \multicolumn{1}{c|}{} & \multicolumn{1}{c|}{} &  \\ \hline
\end{tabular}
\end{center}
\vspace{-0.4cm}
\end{table*}

\subsection{Benchmarks Setup}

\quad Our HoloVIC dataset contains 100k frames of synchronized data, with 70\% dedicated to holographic intersection data and 30\% allocated for vehicle-infrastructure cooperation (VIC) data. To ensure privacy, we have applied blurring to all faces, vehicle plates and road signs in all data. The details of the dataset are illustrated in Tab.\ref{tab3} The holographic intersection data is collected from five different intersections with four distinct sensor layouts. The VIC data collection begins when a vehicle enters an intersection and ends when it exits the corresponding intersection area. 

HoloVIC is divided into training, testing, and validation sets, with a ratio of 50\%, 40\%, and 10\% respectively. The training and validation sets include ground truth labels, while the testing set only provides data. One of the five holographic intersections is exclusively assigned to the testing set, while the remaining intersections are included in all three sets. Algorithms can be submitted to our benchmark for online evaluation of the corresponding task.

\subsection{Sensor Specifications}
\quad The detailed specifications of all devices as shown in Tab.\ref{tab2}. Due to the varying distances between the areas and poles, we select Cameras with different focal lengths and Field of View (FOV) to cover the areas. All devices are synchronized in time via Network Time Protocol (NTP) before data collection, we utilize a time interval of 100ms as the global timestamp for intersections, and match the frame from each device with the nearest timestamp adjacent to the global timestamp. This process ultimately yields synchronized multi-sensor data at a frame rate of 10 fps.

\subsection{Baselines}
\subsubsection{3D Detection}

\quad \textbf{Monocular 3D Detection} To evaluate the performance of image-only monocular 3D detection, we selected widely-used methods as our Mono3D baseline models FCOS3D~\cite{wang2021fcos3d} for validation, which is a fully convolutional single-stage detector and transforms 7-DoF 3D targets to the image domain and decouple them as 2D and 3D attributes. In addition, we choose PGD~\cite{wang2022probabilistic} as a comparison to our baseline method.

\textbf{Lidar 3D Detection}  To demonstrate the capabilities of well-known Lidar Detectors in our Lidar 3D task, we implemented four methods with different architectures: SECOND~\cite{yan2018second}, Pointpillars~\cite{lang2019pointpillars}, PVRCNN~\cite{shi2020pv}, and Transfusion-L~\cite{bai2022transfusion}, the methods utilized different tech-niques such as Voxelization, Pillar-based, Two-stage Re-gion Proposals, and Transformers to achieve accurate and efficient detection in pointcloud, where we select Pointpillars as our baseline for Lidar 3D Detection.

\subsubsection{Multiple Object Tracking}

\quad \textbf{2D/3D MOT} We follow the Tracking-by-Detection paradigm using 2D or 3D bounding boxes as inputs. The 2D boxes are generated by the Yolov8~\cite{reis2023real} detector pretrained on the COCO dataset~\cite{lin2014microsoft}. We selected DeepSort~\cite{wojke2017simple}, Tracktor~\cite{bergmann2019tracking}, and FairMOT~\cite{zhang2021fairmot} as our 2D MOT baseline. And the 3D boxes from image and pointclouds are provided by FCOS3D and Pointpillars respectively. As for the 3D tracker, we select AB3DMOT~\cite{2020AB3DMOT} as 3D MOT baseline.

\subsubsection{Multi-Sensor Tracking}
\quad \textbf{Multi-Sensor Multi-Object Tracking} We have develop a multi-sensor late-fusion framework to fuse the local tracklets as global trajectories based on the intersection coordinate system. In this framework, we utilize the 3D MOT baselines to generate tracking results, which serve as inputs for the fusion process. To associate the same objects from different Cameras, we employ the Hungarian Algorithm~\cite{kuhn1955hungarian} to solve the bipartite graph problem. Finally, we integrates the local information from each device to generate global trajectory information.

\subsubsection{VIC Perception}
\quad \textbf{Vehicle-Intersection Cooperative Perception} focuses on fusing both data and perception results from road and vehicle sides. We design a VIC late-fusion framework as the baseline. Both sides generate perception results as inputs for late-fusion. The 3D boxes from the road-side perception results are transformed to the ego-vehicle coordinate system based on the relative positions between coordinates. We utilize the Hungarian Algorithm~\cite{kuhn1955hungarian} to match the 3D boxes belonging to the same targets on both sides. The 3D bounding boxes are associated in consecutive frames using the AB3DMOT algorithm to generate the fused trajectory between the vehicle and the road.

\subsection{Analysis}

\quad\   \textbf{Performance of Single-Sensor Perception,} we evaluated the perception results on single sensor in the HoloVIC validation set. The ground truth was defined by covering objects visible in both the Camera and Lidar. The detection results based on Mono3D and Lidar 3D are shown in Tab.~\ref{tab4}, where we evaluated 3D AP, BEV AP, and AOS separately for three different categories. Lidar 3D generally higher than Mono3D for AP, especially for cyclists and pedestrians. However, in terms of orientation error, the AOS metric is generally higher for image-based perception compared to Lidar 3D. Regarding tracking, as shown in Tab.~\ref{tab5}, Mono3D-based tracking exhibits relatively poorer performance due to weaker detection results compared to other methods. Lidar 3D tracking is performed in 3D space, which mitigates the impact of occlusion between objects that often leads to tracklet mixing in 2D tracking. However, AB3DMOT~\cite{2020AB3DMOT} only considers the position and motion of the target for tracking and cannot reconnect lost targets through strategies like ReID used in 2D MOT. Therefore, the MOTA and IDF1 metrics fall between the performance of 2D MOT methods.

\begin{figure}
\vspace{-0.0cm}
	\centering
	\includegraphics[width=7.5cm]{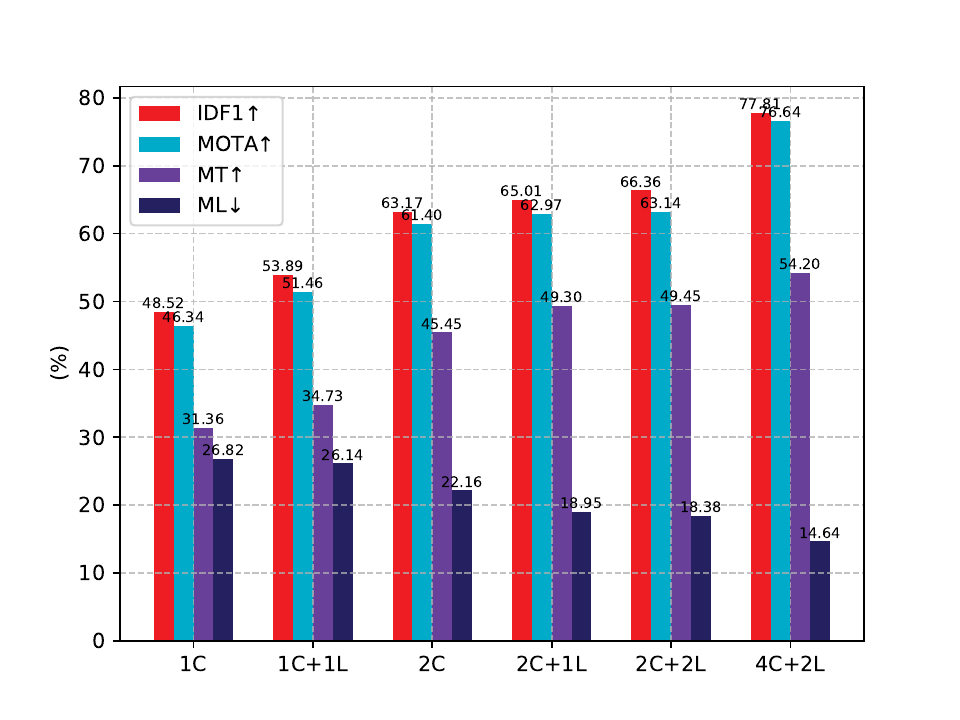}
 \vspace{-0.2cm}
	\caption{The impact of different sensor-layouts on multi-sensor tracking accuracy in Int-1 validation set.}
	\label{fig_exp4}
\vspace{-0.4cm}
\end{figure}

\textbf{Performance with different Sensor-layouts,} we evaluated the performance using the Int-1 (4C+2L) validation set. The ground truth data was obtained by targets within the areas covered by the field of view of all devices. As depicted in  Fig.\ref{fig_exp4}, we evaluate IDF1, MOTA, MT, and ML in the intersection coordinate system based on six different sensor layouts. As the number of sensors increased, all metrics showed gradual improvement. Multiple viewpoints provided better coverage of blind spots and enhanced trajectory continuity. Therefore, the 2C method achieved a higher improvement in the 1C results compared to 1C+1L, with an increase of 9.28\%, 9.94\%, 10.72\%, and 3.98\% respectively. Similarly, 4C+2L exhibited improvements of 11.36\%, 13.5\%, 4.75\%, 3.74\% compared to 2C+2L.

\textbf{Performance from VIC,} we evaluated the detection and tracking performance within ranges of 30m, 50m, and 70m around the center of the ego-vehicle. The vehicle travels within a range of 70m from the intersection coordinate system. The ground truth includes all bounding boxes on both the road-side and vehicle-side within the corresponding evaluation range. We separately evaluated the detection and tracking performance of only the vehicle-side perception, VIC perception with 1C+1L, and 4C+1L based on Scene of Int-1. As shown in Tab.\ref{tab6}, the performance of VIC perception is better than vehicle-side in all metrics. Moreover, increasing the number of sensors from the roadside can greatly improve the performance for the vehicle. The perception capability beyond 50m deteriorates rapidly due to target occlusion and the limited sensing range of the ego-vehicle sensors. Through VIC, the performance is improved up to 38.8\% and 40.76\% in terms of AP and AOS, respectively. The accuracy in the 50-70m range on the vehicle-side remains consistent with the range of 0-50m. Additionally, roadside perception provides an 8.9\% and 14.55\% improvement in IDF1 and MOTA for the vehicle-side.

\section{Conclusion}

\quad In this paper, we constructed several holographic intersections with three different types of sensors and four different sensor-layouts. We annotated 11.47M 3D Boxes in total on 100k synchronous frames from different sensors. We propose a large-scale holographic intersection and vehicle-infrastructure cooperative dataset, HoloVIC. Furthermore, the dataset is divided into multiple tasks in various dimensions for further research on perception models. In the future, we plan to expand more tasks and benchmarks based on HoloVIC, such as trajectory prediction, and explore the additional benefits that roadside perception can bring to vehicle-side perception.

%% file: sec/X_suppl.tex
\clearpage
\maketitlesupplementary
\appendix
\setcounter{page}{1}
\setcounter{section}{0}
\setcounter{figure}{0}
\setcounter{equation}{0}
\renewcommand{\thefigure}{A\arabic{figure}}

\begin{figure*}
	\centering
	\includegraphics[width=17cm]{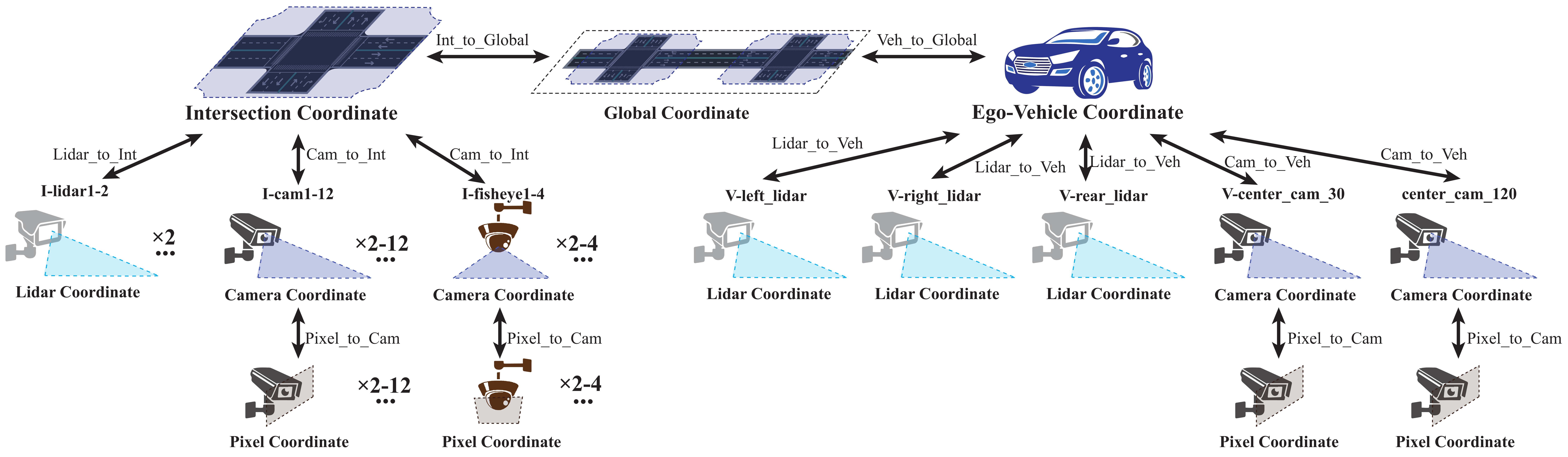}
	\caption{The transformation relationships between all coordinate systems in HoloVIC.}
	\label{figA1}
\vspace{-0.3cm}
\end{figure*}

\section{Coordinates Transformation}
\quad In Sec.3.2, we introduced all the coordinate systems involved in HoloVIC, as shown in Fig.\ref{fig3}. The coordinate systems include the Lidar Coordinate $(l_x,l_y,l_z)$, Camera Coordinate $(c_x,c_y,c_z)$, and Pixel Coordinate $(u,v)$, which represent the positions in their respective sensors. The Intersection Coordinate $(\sigma_x,\sigma_y,\sigma_z)$ and Ego-Vehicle Coordinate $(\delta_x,\delta_y,\delta_z)$ are used for unifying the coordinate systems of their respective sensors, and the Global Coordinate $(\omega_x,\omega_y,\omega_z)$ mainly aligns the Intersection Coordinate and Vehicle Coordinate. The transformation relationships between all coordinate systems are shown in Fig.\ref{figA1}, where  which include both forward and inverse transformations for five processes.

\subsection{Lidar$\Leftrightarrow$Intersection/Vehicle}

\quad\textbf{Intersection to Lidar:} Given a point in the Intersection Coordinate: $(\sigma_x,\sigma_y,\sigma_z)$, the transformation of this point from the Intersection Coordinate System to the Lidar Coordinate $(l_x,l_y,l_z)$ is defined as follows:

\begin{eqnarray}
\label{aeq1}
\begin{pmatrix}
l_x\\
l_y\\
l_z \\
1
\end{pmatrix}=
{RT}_{I2L}
\begin{pmatrix}
\sigma_x\\
\sigma_y\\
\sigma_z \\
1
\end{pmatrix},
\end{eqnarray}
where ${RT}_{I2L}\in\mathbb{R}^{4\times4}$ is a Rotational Translation of a homogeneous matrix for Intersection Coordinate to Lidar Coordinate, which is formulated as:

\begin{eqnarray}
\label{aeq2}
{RT}_{I2L}^{4\times4}=
\begin{bmatrix}
{R}_L^{3\times3}&{T}_L^{3\times1}\\ 
0&1
\end{bmatrix}
\end{eqnarray}

\textbf{Lidar to Intersection:} Given a point in Lidar Coordinate $(l_x,l_y,l_z)$, the transformation of the point to Intersection Coordinate $(\sigma_x,\sigma_y,\sigma_z)$ is defined as:

\begin{eqnarray}
\label{aeq3}
\begin{pmatrix}
\sigma_x\\
\sigma_y\\
\sigma_z \\
1
\end{pmatrix}=
{RT}_{L2I}
\begin{pmatrix}
l_x\\
l_y\\
l_z \\
1
\end{pmatrix},\ \ {RT}_{L2I}={RT}_{I2L}^{-1}
\label{supeq3}
\end{eqnarray}
where ${RT}_{L2I}\in\mathbb{R}^{4\times4}$ is the inverse of ${RT}_{I2L}\in\mathbb{R}^{4\times4}$.

\textbf{Vehicle to Lidar:} Similar to Eq.(\ref{aeq1}) and Eq.(\ref{aeq2}), a point $(\delta_x,\delta_y,\delta_z)$ in the Vehicle Coordinate is transformed by using Eq.(\ref{aeq1}) through the ${RT}_{V2L}\in\mathbb{R}^{4\times4}$ to obtain the point in the Lidar Coordinate of Vehicle $(l_x,l_y,l_z)$.

\textbf{Lidar to Vehicle:} Similar to Eq.(\ref{aeq3}), a point $(l_x,l_y,l_z)$ in the Lidar Coordinate of Vehicle is transformed by using Eq.(\ref{aeq3}) through the ${RT}_{L2V}\in\mathbb{R}^{4\times4}$ to obtain the point in the Vehicle Coordinate $(\delta_x,\delta_y,\delta_z)$, where ${RT}_{L2V}$ is the inverse of ${RT}_{V2L}$.

\subsection{Camera$\Leftrightarrow$Intersection/Vehicle}

\quad\textbf{Intersection to Camera:} Given a point in Intersection Coordinate $(\sigma_x,\sigma_y,\sigma_z)$, the transformation of the point to Camera Coordinate $(c_x,c_y,c_z)$ is defined as:

\begin{eqnarray}
\label{aeq4}
\begin{pmatrix}
c_x\\
c_y\\
c_z \\
1
\end{pmatrix}=
S^{4\times4}
{RT}_{I2C}
\begin{pmatrix}
\sigma_x\\
\sigma_y\\
\sigma_z \\
1
\end{pmatrix}
\end{eqnarray}
where ${RT}_{I2C}\in\mathbb{R}^{4\times4}$ is a Rotational Translation of a homogeneous matrix for Intersection Coordinate to Camera Coordinate, and $S$ is utilized for mapping coordinate axes ($X\rightarrow Z,Y\rightarrow -X, Z\rightarrow -Y$), which are formulated as:
\begin{eqnarray}
\label{aeq5}
{RT}_{I2C}^{4\times4}=
\begin{bmatrix}
{R}_C^{3\times3}&{T}_C^{3\times1}\\ 
0&1
\end{bmatrix},\ \ 
S^{4\times4}=
\begin{bmatrix}
0&-1&0&0\\
0&0&-1&0\\
1&0&0&0\\
0&0&0&1\\
\end{bmatrix}
\end{eqnarray}

In addition, we define the ground plane vector $\phi_C^{1\times4}$ and $\phi_I^{1\times4}$ relative to the Camera Coordinate and Intersection Coordinate, respectively. And we define the the ground plane in Intersection is $Z=0$, and then the $\phi_I$ is $[0,0,1,0]$. The points set in ground plane of Intersection Coordinate $\{p:(p_x,p_y,p_z), p_z=0\}$ and the points set in ground plane of Camera Coordinate $\{q:(q_x,q_y,q_z), q_z=0\}$ satisfy:

\begin{eqnarray}
\label{aeq6}
\phi_I^{1\times4}
\begin{pmatrix}
p_x\\
p_y\\
p_z \\
1\\
\end{pmatrix}
=0,\quad
\phi_C^{1\times4}
\begin{pmatrix}
q_x\\
q_y\\
q_z \\
1\\
\end{pmatrix}
=0
\end{eqnarray}

The points in Intersection Coordinate ${p:(p_x,p_y,p_z)}$ are transformed to Camera Coordinate ${q:(q_x,q_y,q_z)}$ by Eq.(\ref{aeq4}): 

\begin{eqnarray}
\label{aeq7}
\begin{pmatrix}
q_x\\
q_y\\
q_z \\
1
\end{pmatrix}=
S^{4\times4}
{RT}_{I2C}
\begin{pmatrix}
p_x\\
p_y\\
p_z \\
1
\end{pmatrix}
\end{eqnarray}
Combining Eq.(\ref{aeq6}) and Eq.(\ref{aeq7}) we obtain:
\begin{eqnarray}
\label{aeq8}
\phi_I^{1\times4}
{[S^{4\times4}{RT}_{I2C}]}^{-1}
\begin{pmatrix}
q_x\\
q_y\\
q_z \\
1\\
\end{pmatrix}
=0
\end{eqnarray}
thus we derive the ground plane vector $\phi_C$ is equal to:
\begin{eqnarray}
\label{aeq9}
\phi_C^{1\times4}=
\phi_I^{1\times4}
{[S^{4\times4}{RT}_{I2C}]}^{-1}
\end{eqnarray}


\textbf{Camera to Intersection:} Given a point $p_c$ in the Camera Coordinate: $(c_x,c_y,c_z)$, the transformation of this point from the Camera Coordinate System to the Intersection Coordinate $(\sigma_x,\sigma_y,\sigma_z)$ is defined as follows:
\begin{eqnarray}
\label{aeq10}
\begin{pmatrix}
\sigma_x\\
\sigma_y\\
\sigma_z \\
1
\end{pmatrix}=
{RT}_{C2I}S^{-1}
\begin{pmatrix}
c_x\\
c_y\\
c_z \\
1
\end{pmatrix},\ \ {RT}_{C2I}={RT}_{I2C}^{-1}
\end{eqnarray}
where ${RT}_{C2I}\in\mathbb{R}^{4\times4}$ is the inverse of ${RT}_{I2C}\in\mathbb{R}^{4\times4}$.

\textbf{Vehicle to Camera:} Similar to Eq.(\ref{aeq4}) and Eq.(\ref{aeq5}), a point $(\delta_x,\delta_y,\delta_z)$ in the Vehicle Coordinate is transformed by using Eq.(\ref{aeq4}) through the ${RT}_{V2C}\in\mathbb{R}^{4\times4}$ to obtain the point in the Camera Coordinate of Vehicle $(c_x,c_y,c_z)$.

\textbf{Camera to Vehicle:} Similar to Eq.(\ref{aeq10}), a point $(c_x,c_y,c_z)$ in the Camera Coordinate of Vehicle is transformed by using Eq.(\ref{aeq10}) through the ${RT}_{C2V}\in\mathbb{R}^{4\times4}$ to obtain the point in the Vehicle Coordinate $(\delta_x,\delta_y,\delta_z)$, where ${RT}_{C2V}$ is the inverse of ${RT}_{V2C}$.

\subsection{Pixel$\Leftrightarrow$Camera}

\quad\textbf{Camera to Pixel:} Given a position point $p_c$ in the camera coordinate system: $(c_x,c_y,c_z)$, the projection of this point from the camera coordinate system to undistorted image in the pixel coordinate system is defined as follows:

\begin{eqnarray}
\label{aeq11}
Z_c
\begin{pmatrix}
u\\
v\\
1
\end{pmatrix}=
K^{3\times3}
\begin{pmatrix}
c_x\\
c_y\\
c_z 
\end{pmatrix}
\end{eqnarray}
where $K\in\mathbb{R}^{3\times3}$ indicates the intrinsic matrix of camera which is calibrated by Chessboard Calibration. $f_x,f_y$ donote the focal of the camera in $x$-axis, $y$-axis. $u_0,v_0$ represent the center of image. $Z_c$ indicates the distance from point $p_c$ to the projection plane of camera.

\textbf{Pixel to Camera:} Since the transformation from the pixel coordinate system to the camera coordinate system is a 2D to 3D process, let's assume that we select the points $(u, v)$ in the image that corresponds to a point on the ground plane in real scene, and it is projected onto the camera coordinate system as $(c_x,c_y,c_z)$. Before the projection, we have to calculate distance between the points on Camera Coordinate to plane of camera $Z_c$, which is calculated as:
\begin{eqnarray}
\label{aeq12}
Z_c=\frac{-d}{(u[a,b,c]K^{-1}_{|0}+v[a,b,c]K^{-1}_{|1}+[a,b,c]K^{-1}_{|2})}
\end{eqnarray}
where $\phi_C\in\mathbb{R}^{1\times4}:[a,b,c,d]$ is the ground plane vector for Camera Coordinate, which is introduced in Sec.A.2. $K^{-1}\in\mathbb{R}^{3\times3}$ is the inverse of camera intrinsic $K$, and in Eq.(\ref{aeq12}), $K^{-1}_{|i}\in\mathbb{R}^{3\times1}$ indicates the $i$-th column of the $K^{-1}$. The transformation from Pixel Coordinate to Camera Coordinate is defined as:
\begin{eqnarray}
\label{aeq13}
\begin{pmatrix}
c_x\\
c_y\\
c_z\\
1
\end{pmatrix}=
{\begin{bmatrix}
{K}^{-1}&0\\ 
0&1
\end{bmatrix}}^{4\times4}
\begin{pmatrix}
Z_c u\\
Z_c v\\
Z_c\\
1
\end{pmatrix}
\end{eqnarray}

\subsection{Global$\Leftrightarrow$Intersection:} 

\quad \textbf{Global to Intersection:} Both of Global Coordinate and Intersection belong to East-North-Up (ENU) Coordinate. Given a point in Global Coordinate $(\omega_x,\omega_y,\omega_z)$, the transformation from Global Coordinate to Intersection Coordinate $(\sigma_x,\sigma_y,\sigma_z)$ is defined as:
\begin{eqnarray}
\begin{pmatrix}
\sigma_x\\
\sigma_y\\
\sigma_z \\
1
\end{pmatrix}=
\begin{bmatrix}
{E}^{3\times3}&{T}_{G2I}^{3\times1}\\ 
0&1
\end{bmatrix}
\begin{pmatrix}
\omega_x\\
\omega_y\\
\omega_z \\
1
\end{pmatrix}
\end{eqnarray}
\begin{eqnarray}
{T}_{G2I}^{3\times1}=
\begin{pmatrix}
\omega_{x0}\\
\omega_{y0}\\
\omega_{z0}
\end{pmatrix}-
\begin{pmatrix}
\sigma_{x0}\\
\sigma_{y0}\\
\sigma_{z0}
\end{pmatrix}
\end{eqnarray}
where $E\in\mathbb{R}^{3\times3}$ is a identity matrix, ${T}_{G2I}^{3\times1}$ indicates the translation matrix between Global Coordinate and Intersection, $(\omega_{x0},\omega_{y0},\omega_{z0})$ and $(\sigma_{x0},\sigma_{y0},\sigma_{z0})$ denote the original of Global and Intersection, respectively.

\textbf{Global to Intersection:} The transformation from Intersection to Global is formulated as:
\begin{eqnarray}
\begin{pmatrix}
\omega_x\\
\omega_y\\
\omega_z \\
1
\end{pmatrix}=
\begin{bmatrix}
{E}^{3\times3}&{T}_{I2G}^{3\times1}\\ 
0&1
\end{bmatrix}
\begin{pmatrix}
\sigma_x\\
\sigma_y\\
\sigma_z \\
1
\end{pmatrix},\ {T}_{I2G}^{3\times1}=-{T}_{G2I}^{3\times1}
\end{eqnarray}

\subsection{Vehicle$\Leftrightarrow$Global} 
\quad\textbf{Global to Vehicle:} Given a point in Global Coordinate $(\omega_x,\omega_y,\omega_z)$, the rotation and translation matrixes are computed according to GPS, orientation and accelerate of the vehicle by RTK and E-Compass. We directly provide the $RT_{G2V}$ matrix from Global Coordinate to Vehicle Coordinate $(\delta_x,\delta_y,\delta_z)$, which is defined as:
\begin{eqnarray}
\begin{pmatrix}
\delta_x\\
\delta_y\\
\delta_z \\
1
\end{pmatrix}=
\begin{bmatrix}
{R}_{G2V}^{3\times3}&{T}_{G2V}^{3\times1}\\ 
0&1
\end{bmatrix}
\begin{pmatrix}
\omega_x\\
\omega_y\\
\omega_z \\
1
\end{pmatrix}
\end{eqnarray}

\textbf{Vehicle to Global:} The transformation from Vehicle Coordinate $(\delta_x,\delta_y,\delta_z)$ to Global Coordinate $(\omega_x,\omega_y,\omega_z)$ is formulated as:
\begin{eqnarray}
\begin{pmatrix}
\omega_x\\
\omega_y\\
\omega_z \\
1
\end{pmatrix}=
RT_{V2G}
\begin{pmatrix}
\delta_x\\
\delta_y\\
\delta_z \\
1
\end{pmatrix},\ \ RT_{V2G}=RT_{G2V}^{-1}
\end{eqnarray}

\section{Annotation Process}
\subsection{Device Time Synchronization}

\quad All devices at each intersection are connected to a switch, and their time synchronization is achieved through NTP (Network Time Protocol), which ensure the time error is less than 5ms. When collecting data, the cameras and fisheyes capture at a frequency of 25Hz, while the LiDAR operates at 10Hz. 

To construct the dataset, we establish timestamp anchors at a frequency of 10Hz along the timeline. We then select the nearest frame from each device around each anchor and package them into frame batches. Each batch contains synchronized data from all devices at that specific time, and is assigned a corresponding frame index.

\subsection{Calibration}
\quad Once the sensors are deployed at each intersection and time synchronization is complete, we need to calibrate all sensors. For cameras and fisheyes, calibration involves determining the distortion, intrinsic, and extrinsic parameters. For LiDARs, only the extrinsic parameters need to be calibrated. These extrinsic parameters establish the transformation relationship between the intersection/vehicle coordinate system and the device coordinate system, as explained in Sec. A. 

To make it easier for researchers to use our dataset, all images captured by the cameras and fisheyes are undistorted. The coordinate transformation is based on undistorted pixel coordinates and the intersection/vehicle coordinate system. Furthermore, the coordinate transformation between devices can be achieved by linking their extrinsic parameters to the intersection/vehicle coordinate system.
\subsection{Global Annotation}
\quad We merge all the point clouds within each frame batch. Using Eq.(\ref{supeq3}) from the supplementary material, we project each individual point cloud onto the intersection coordinate system. The two sets of point clouds are concatenated together. And then, we annotate the 3D boxes for the targets that appear in the concatenated point cloud scene. This annotation process involves determining the positions, orientations, and categories of the 3D boxes.

Afterwards, the annotated 3D boxes are projected onto the images using the corresponding extrinsic parameters for each camera and fisheye. Annotators are then tasked with performing supplementary annotations for each camera, especially in cases where the target's point cloud is occluded or extends beyond the range captured by the LiDAR, resulting in missed annotations. The annotated boxes are subsequently projected onto the intersection/vehicle coordinate system. However, it is important to note that due to calibration errors, there may be a minuscule number of boxes in the dataset that have inaccurate projections onto the intersection position.

Annotators associate a global ID with the annotated 3D boxes in the timeline. Subsequently, all the 3D boxes are reprojected onto all the devices. Annotators then determine the visibility of each box, indicating which devices can see the box. For example, if a box is only visible in Lidar-2, Camera-1, Camera-3, and Fisheye-1, then the visibility information for that box will have "True" ("Visible") for those devices and "False" ("Invisible") for the rest. There are several situations where a box may be marked as "Invisible": If the global 3D box is outside the field of view of a device; If the object is occluded by other objects that exceeds more than 80\%; If the object appears too small in the image (far from the device). 

\begin{figure*}[t]
	\centering
	\includegraphics[width=17cm]{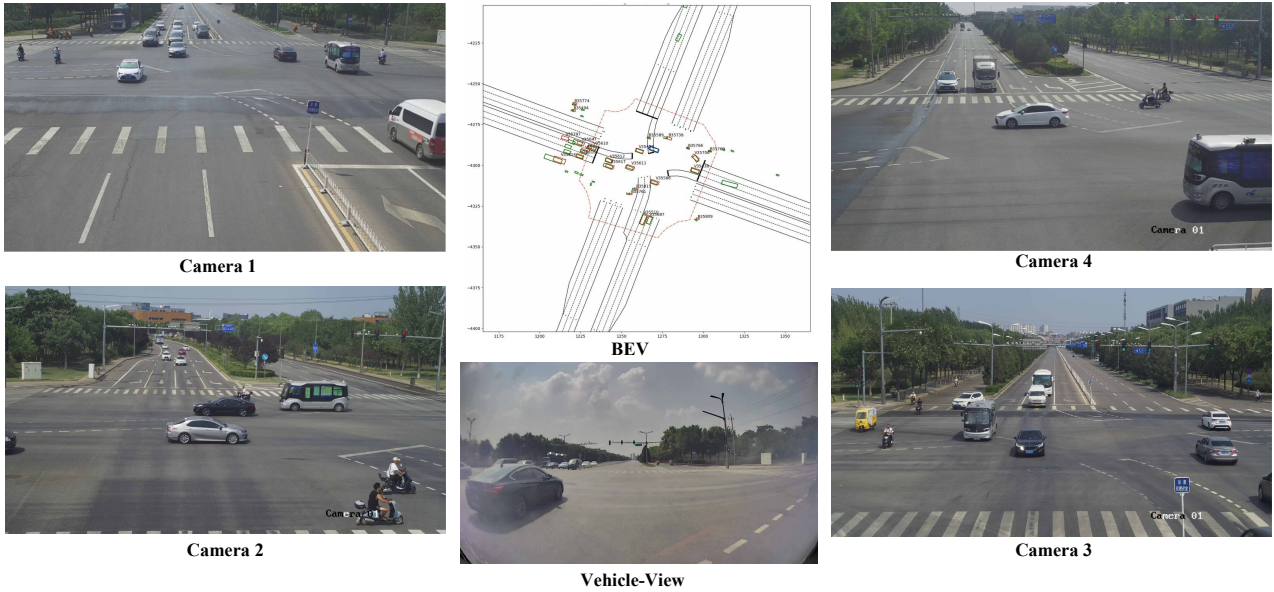}
	\caption{The illustration of Vehicle-Infrastructure Cooperative in VIC-1 at the $4590$-th frame. In the BEV view, the blue rectangular box indicates the ego-vehicle position, the red and green boxes indicate the targets from the Vehicle Coordinate and Intersection Coordinate, respectively.}
	\label{figA6}
\end{figure*}

Considering that many researchers often use trajectory inpainting based on temporal to handle occlusions in tracking tasks, it is worth noting that even though an object may be occluded and not visible in a specific frame, the inpainted boxes can still accurately outline its correct position. Therefore, during evaluation, all of the boxes are labels as "Invisible" are not counted as false positives or false negatives for calculating mAP, MOTA, IDF1, etc., regardless of whether the prediction boxes provided or not.

Both the vehicle-side and the road-side undergo the calibration process described above. Afterwards, all the 3D boxes are transformed onto the global coordinate system. The global ID association is then determined based on the Intersection over Union (IOU) between the 3D boxes of the vehicles and the road.

\section{Visualization of HoloVIC}

\begin{figure}
	\centering
	\includegraphics[width=8.5cm]{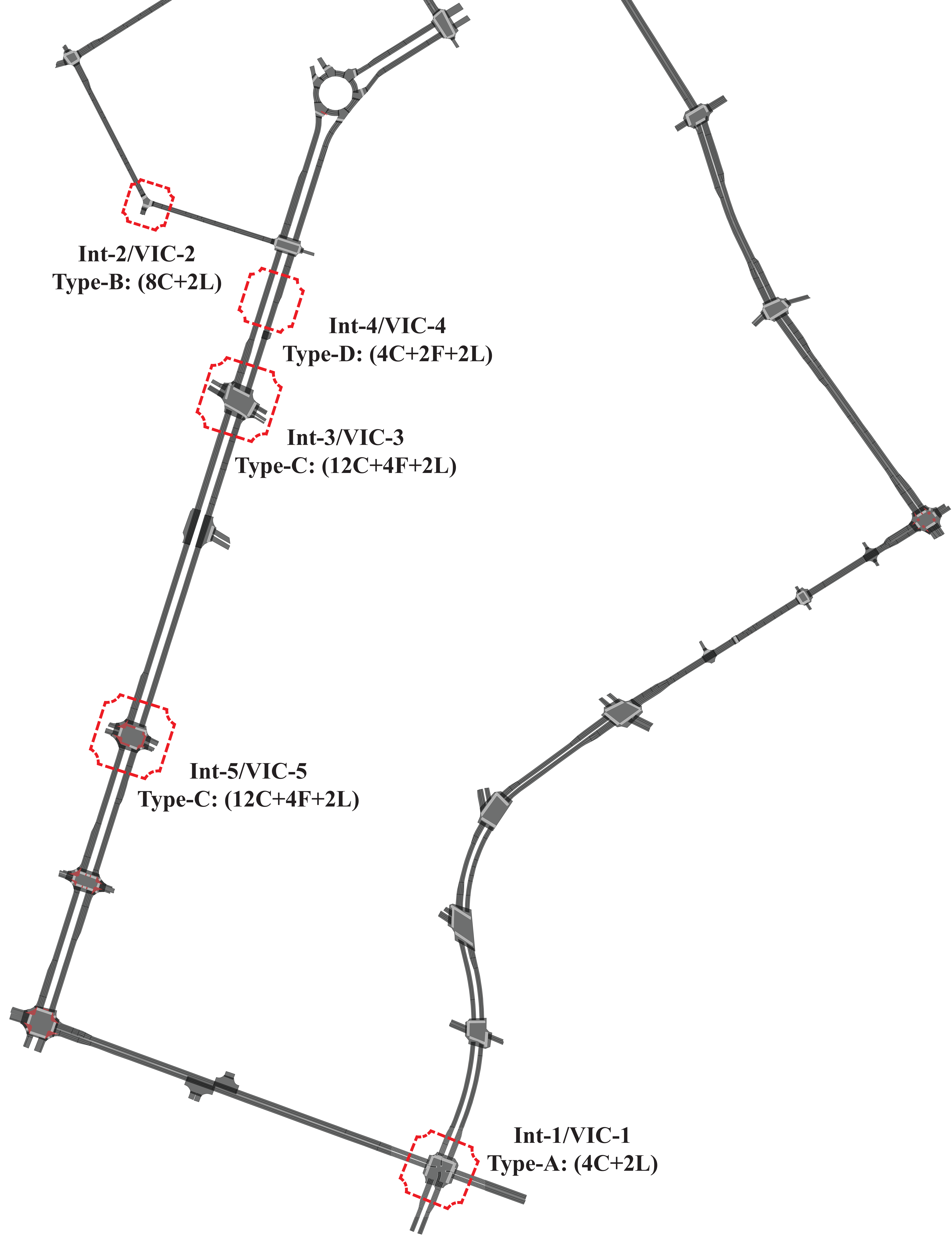}
	\caption{The distribution of all intersections in the HoloVIC Dataset in HD-Map}
	\label{fig4}
\vspace{-0.3cm}
\end{figure}

\begin{figure*}
	\centering
	\includegraphics[width=17cm]{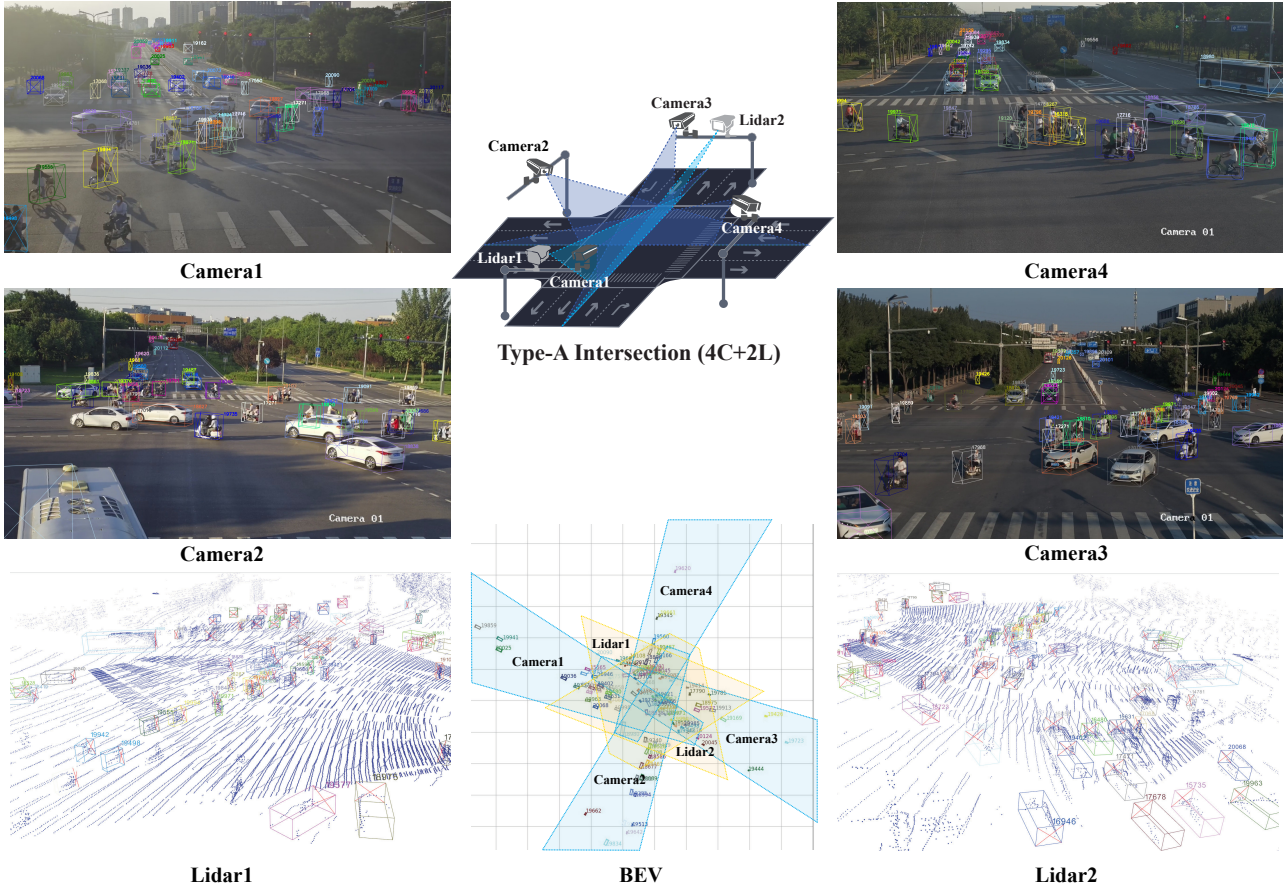}
	\caption{The illustration of Type-A intersection (4C+2L) in Int-1 at the $376$-th frame.}
	\label{figA2}
\end{figure*}

\begin{figure*}
	\centering
	\includegraphics[width=17cm]{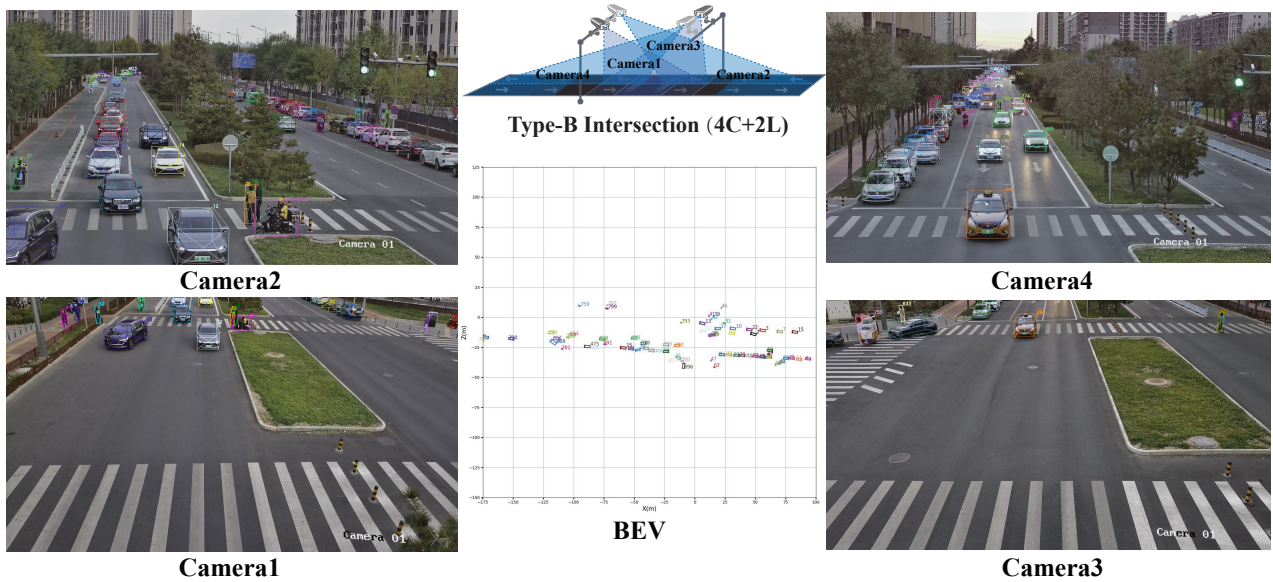}
	\caption{The illustration of Type-B intersection with two opposite viewpoints (4C+2L) in Int-2 at the $754$-th frame.}
	\label{figA3}
\end{figure*}

\begin{figure*}
	\centering
	\includegraphics[width=17cm]{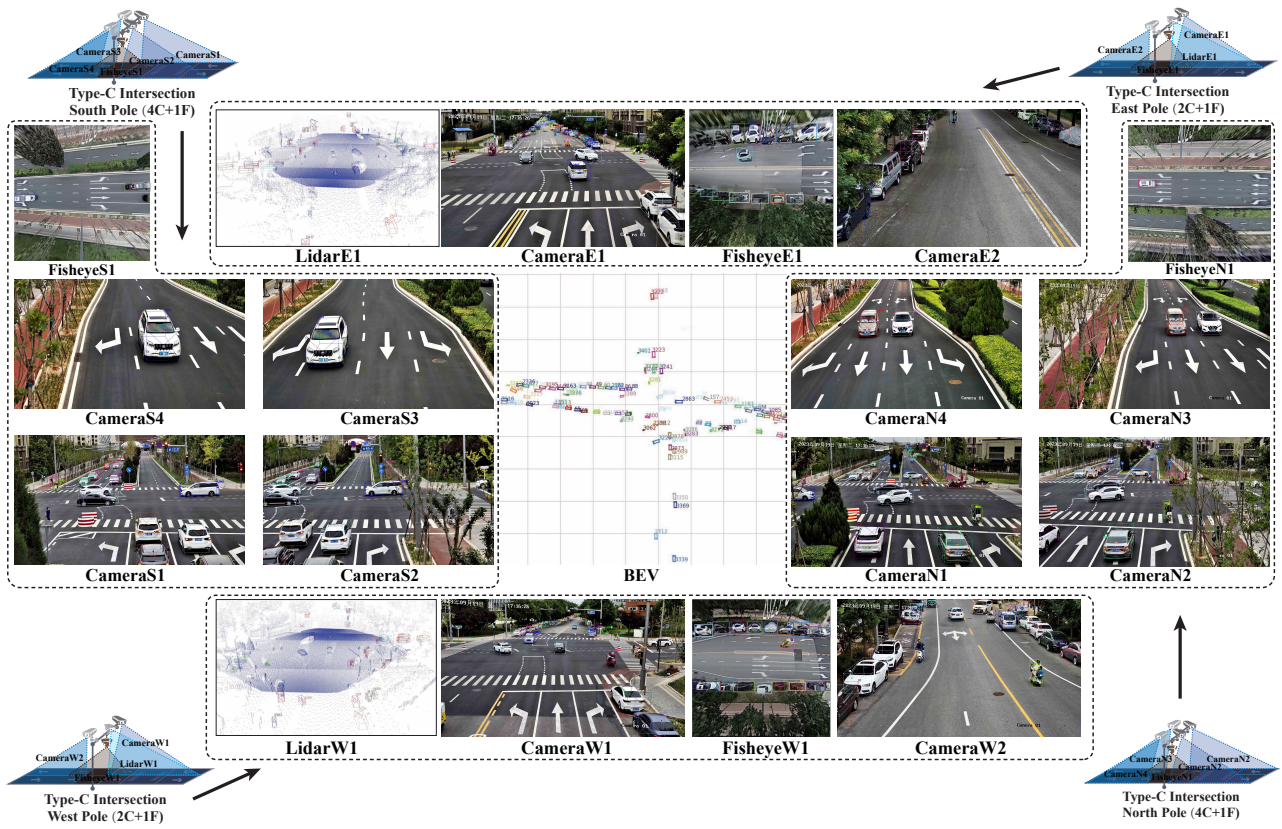}
    \vspace{-0.3cm}
	\caption{The illustration of Type-C intersection (12C+4F+2L) in Int-3 at the $1105$-th frame.}
	\label{figA5}
\end{figure*}

\begin{figure*}
	\centering
	\includegraphics[width=17cm]{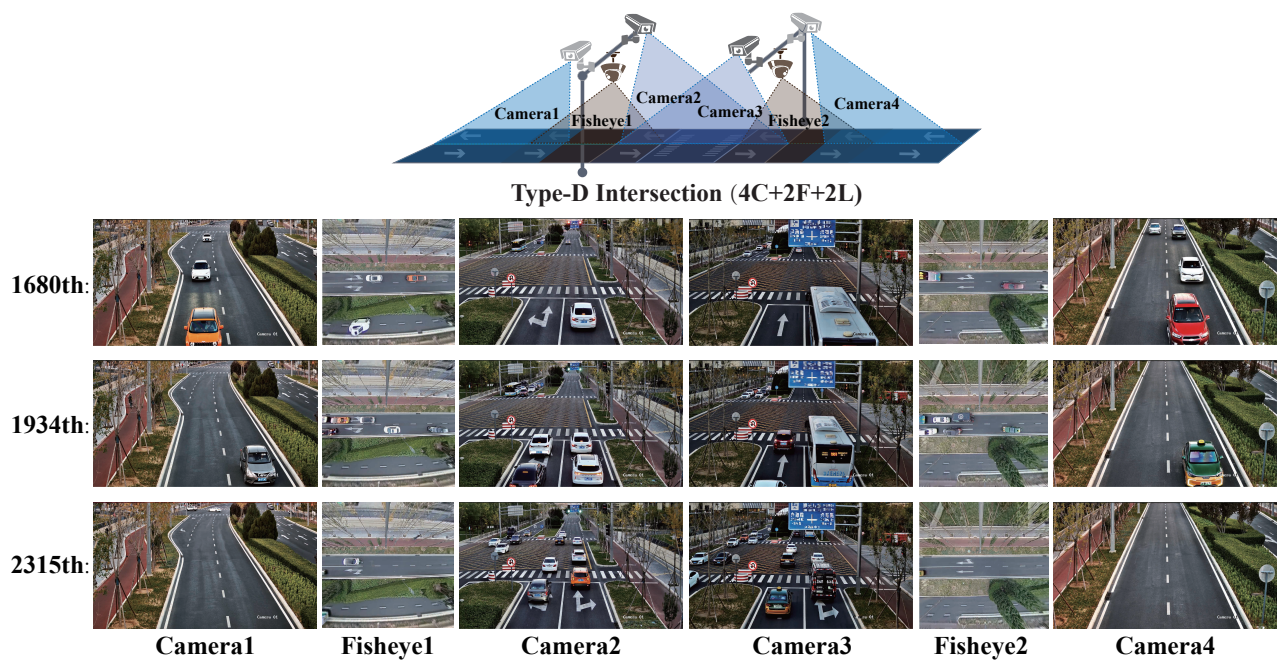}
	\caption{The illustration of Type-D intersection (4C+2F+2L) in Int-4 at the $1680$-th, $1934$-th and $2315$-th frame.}
	\label{figA4}
\end{figure*}

\quad We show all of the visualization results involving all of intersections Int-1/VIC-1 to Int-5/VIC-5. The distribution of all intersections in the HoloVIC Dataset in HD-Map is shown in Fig.\ref{fig4}. The red dashed box identifies the corresponding intersection number for each intersection (Int-1/VIC-1)-(Int-5/VIC-5), which share the same Global Coordinate System. The illustration of Vehicle-Infrastructure Cooperative is illustrated in Fig.\ref{figA6}. The illustration of intersections with different sensor layouts (Type A-D) are shown in Fig.\ref{figA2}-\ref{figA4}.

